\documentclass[11pt]{article}

\usepackage[preprint]{acl}
\usepackage{amsmath} 
\usepackage{times}
\usepackage{latexsym}
\usepackage{arydshln} 

\usepackage[T1]{fontenc}
\usepackage[utf8]{inputenc}

\usepackage{microtype}

\usepackage{inconsolata}

\usepackage{graphicx}

\usepackage{booktabs}
\usepackage{multirow}
\usepackage{graphicx}
\usepackage[table]{xcolor}
\usepackage{geometry}
\definecolor{applegreen}{rgb}{0.55, 0.71, 0.0}
\definecolor{heatred}{RGB}{220, 20, 60} % Custom RGB
\definecolor{heatblue}{HTML}{4080BF}
\usepackage[utf8]{inputenc}
\usepackage{xcolor}
\usepackage{tcolorbox}
\usepackage{geometry}

\usepackage{tabularx}
\usepackage{adjustbox}
\usepackage{multirow}

\usepackage[utf8]{inputenc}
\usepackage{algorithm}
\usepackage{algorithmic}
\usepackage{amsmath}
\usepackage{amssymb} 

\usepackage{listings}
\usepackage{xcolor}
\usepackage{tcolorbox}
\tcbuselibrary{breakable} % Add this line
\usepackage{tcolorbox}
\tcbuselibrary{skins}

\newcommand{\eg}{\textit{e.g.}}

\title{TaskPress: Query-Agnostic KV Cache Compression via \\ Task-Guided Pruning}

\author{
  Wonpyo Park$^{1}$ \quad Seung-won Hwang$^{2}$ \\
  \vspace{0.2cm} % 저자와 소속 사이의 간격을 살짝 벌려줌
  \hspace*{1.1cm} $^1$Google \quad \hspace{.1cm} $^2$Seoul National University
}

\begin{document}
\maketitle
\begin{abstract}

Long-context inference with large language models (LLMs) is constrained by the linear growth of the key–value (KV) cache to sequence length. 
While pruning offers mitigation, prevailing methods determine query-specific token importance that cannot be reused across unseen queries. In contrast, we introduce TaskPress, a framework for task-guided, query-agnostic KV cache eviction. 
Instead of optimizing the cache for a single query, TaskPress constructs a reusable memory representation conditioned on a high-level task guide. The guide functions as a "meta-query" during prefill to filter irrelevant tokens before downstream queries are issued.
In addition, TaskPress leverages quantization scale factors as a zero-cost signal for detecting influential representation outliers, providing an efficient proxy for token importance.
 Experiments conducted on various tasks with long context input demonstrate that TaskPress efficiently creates a compact, reusable cache across diverse queries. 
 
 %Our code is available at: \href{https://anonymous.4open.science/r/taskpress-8C66/}{https://anonymous.4open.science/r/taskpress}

% Large Language Models (LLMs) encounter severe memory and latency bottlenecks in long-context scenarios due to the linear growth of the Key-Value (KV) cache. While token pruning can alleviate this, prevailing methods rely on query-dependent importance, tailoring the cache to a specific user input. This paradigm renders context reuse inefficient in multi-turn applications, such as agents or RAG, as the compression must be re-executed for each new query. To address this, we introduce TaskPress, a task-guided, query-agnostic eviction framework. Our approach is built on the intuition that the task guide serves as a representative query, encapsulating the semantic scope of potential future interactions. By leveraging the cross-attention map between this representative query and the input context, TaskPress proactively filters task-irrelevant tokens during the prefill phase. Furthermore, we augment this attention-based selection with a novel, zero-cost importance metric: given that quantization is a standard practice in modern inference, we repurpose the pre-existing quantization scale factors to identify critical outlier tokens within Value states. Extensive evaluations on LongBench and RULER demonstrate that TaskPress significantly outperforms existing query-agnostic baselines, maintaining high accuracy while achieving substantial memory reduction.

\end{abstract}

\section{Introduction}

Large language models (LLMs) increasingly operate over long contexts. enabling novel applications reaching up to 2M tokens \cite{comanici2025gemini}. 
During inference, transformers cache intermediate key and value (KV) representations, which 
incurs significant memory overhead due to its linear growth with respect to sequence length.
As a result, KV cache often exceeds model weights, saturating memory and degrading throughput.

\begin{figure}[h]
    \centering
    \begin{tcolorbox}[colback=white, colframe=black, title=\textbf{}]
        
        % --- 1. The Inputs ---
        \small \textbf{Query-A:} \textit{\textcolor{heatblue}{What did Lady Eleanor hide?}}
        \vspace{0.2cm}
        \hrule
        \vspace{0.2cm}
        \small \textbf{Query-B:} \textit{\textcolor{heatred}{Did Doctor Black argue near the library?}}
        \vspace{0.2cm}
        \hrule
        \vspace{0.2cm}
        \small \textbf{Task Guide (Meta-Query):} \textit{\textcolor{applegreen}{Track physical evidence, suspicious behaviors, and character alibis.}}
        
        \vspace{0.2cm}
        \hrule
        \vspace{0.2cm}
        
        % --- 2. The Heatmap Content ---
        % Note: \faded is noise. \heat is semantic match. \outlier is critical entity.
        \noindent
        \small \textbf{Context:} The rain lashed violently against the manor windows, drowning out the sound of the wind. It was a cold, miserable Tuesday.

        % \vspace{0.2cm}

        \textcolor{heatblue}{\colorbox{applegreen!30}{Lady Eleanor quietly slipped the arsenic bottle into} \colorbox{applegreen!30}{her velvet purse.}}
        The fireplace crackled warmly in the corner, casting long shadows across the Persian rug. The butler served Earl Grey tea to the guests.

        % \vspace{0.2cm}

        \textcolor{heatred}{\colorbox{applegreen!30}{Doctor Black was seen arguing with the victim near} \colorbox{applegreen!30}{the library just before midnight.}}
        
    \end{tcolorbox}
    \vspace{-0.3cm}

    \caption{Comparison of attended context between specific queries and task guide. Highlighted colors show that conditioning on specific queries restricts attention to unique answers, creating an `overfitted' context when KV cache is pruned in query-dependent manner, \eg, \textcolor{heatblue}{Query-A} fails to generalize to \textcolor{heatred}{Query-B} or vice versa. In contrast, \colorbox{applegreen!30}{Task Guide} acts as a meta-query with a broader semantic scope. Task-level information preserved as compressed KV cache, is optimized to cover  both queries  and remain robust for future inputs.}
    \label{fig:heatmap}
    \vspace{-0.5cm}
\end{figure}

% \textbf{Task-Guided Cache Pruning.} 
%     Visualization of the \textit{TaskPress} selection process. Gray text indicates tokens with low attention scores relative to the Task Guide (Atmosphere/filler). Red highlights indicate high semantic relevance (Evidence/Actions). Bolded tokens (e.g., \textbf{Lady Eleanor}) represent outlier tokens identified in the Value states, which are preserved regardless of attention score to maintain entity consistency.

\begin{figure*}
    \centering
\includegraphics[width=1\textwidth]{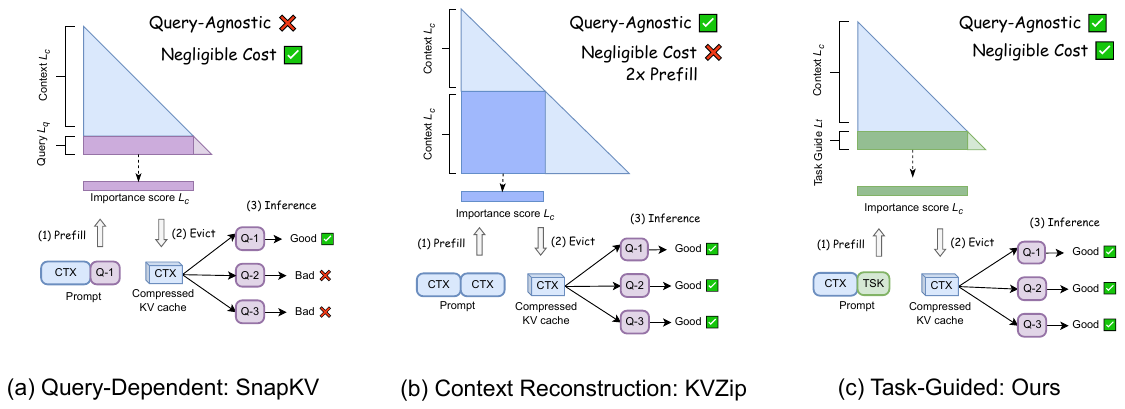}
    \vspace{-173mm}
    \caption{% Comparison of eviction process of each approach. The upper panel shows the attention maps used for importance scoring for each method. The lower panel shows the specific prompt used and the adaptability of the compressed KV cache on other queries. While query-dependent methods \cite{li2024snapkv,tang2024quest,cai2024pyramidkv} are efficient, they lack the flexibility to adapt to other queries. Conversely, context reconstruction methods \cite{kim2025kvzip} accommodate new queries but suffer from significant overhead. TaskPress leverages an evolved task guide (TSK) rather than the query (Q) to compress the context (CTX), achieving a method that is both query-agnostic and structurally robust.
    Comparison of KV cache eviction processes. The upper and lower panels illustrates the attention maps for importance scoring and the resulting cross-query adaptability, respectively. While query-dependent methods \cite{li2024snapkv,tang2024quest,cai2024pyramidkv} lack adaptability and context reconstruction \cite{kim2025kvzip} incurs significant overhead, TaskPress compresses the context (CTX) using a task guide (TSK) rather than a specific query (Q). This ensures a query-agnostic compression with negligible cost.
    }
    \label{fig:method_comparison}
\vspace{-3mm}
\end{figure*}

% While quantization \cite{liu2024spinquant} and KV cache pruning \cite{zhang2023h2o} offer mitigation, 
% existing approaches \cite{li2024snapkv,kim2025kvzip} 
% typically estimate token importance using attention statistics observed during decoding and dynamically evict those less relevant, 
% as shown in Fig \ref{fig:method_comparison} (a). 
% Although effective for single-query inference, these methods remain inherently \emph{query-dependent}: token importance is determined only after a specific query has been observed, such that
% cache cannot be reused for new query~\cite{li2024snapkv}. Conversely, context reconstruction methods \cite{kim2025kvzip} accommodate new queries but suffer from significant overhead, resulting in a $2\times$ increase in prompt length as illustrated in Fig \ref{fig:method_comparison} (b). This highlights an need for efficient query-agnostic compression strategies.

Although quantization \cite{liu2024spinquant} and KV cache pruning \cite{zhang2023h2o} reduce memory costs, existing approaches \cite{li2024snapkv,kim2025kvzip} typically rely on decoding attention to dynamically evict less relevant tokens (Fig \ref{fig:method_comparison}a). This makes them inherently \emph{query-dependent}: because token importance relies on a specific query, the compressed cache cannot be reused for new queries \cite{li2024snapkv}. Alternatively, context reconstruction methods \cite{kim2025kvzip} support multiple queries but introduce significant overhead, doubling the prompt length (Fig \ref{fig:method_comparison}b). These limitations underscore the need for efficient, \emph{query-agnostic} compression strategies.

However, many practical workloads build on tasks, serving  multiple queries. For example, Fig 1 shows that each query attends to different context therefore constructing query-specific KV caches leads to redundant computation and prevents efficient reuse of the compressed context.

In this work, we propose a different perspective: instead of optimizing KV pruning for a single query, we aim to construct a \emph{query-agnostic compressed memory} that can serve future queries drawn from the same task. 
% Our key insight is use \emph{task guide} as a fixed comprehensive queries encompass potential future queries of the given task.
% We aim to construct a \emph{query-agnostic compressed memory} guided by a \emph{task guide} that encapsulates potential future queries over the same context.
By doing so, we move KV pruning from query-time to task-time. 
This decoupling allows the compressed cache to remain highly reusable across multiple turns and robust even when subsequent user queries exhibit moderate task drifts from the original intent.

% The task guide should act as a ``meta-query'' defining the semantic scope to be preserved. 
% For example, in a question answering task over a lengthy financial report, rather than pruning the KV cache based on a narrow, specific query like ``What was the total revenue in Q3?'', a task guide might instruct the model to ``Identify and extract all financial metrics, dates, and key performance indicators.'' This broader directive ensures that the compressed cache retains sufficient relevant information to answer not only the initial question but also subsequent queries about operating costs or net income.

To this end, we realize this conceptual shift through \textbf{TaskPress}, a framework for task-guided, query-agnostic KV cache eviction. 
To construct this guide, we propose a dual approach depending on the deployment scenario. For \emph{known tasks}, we find that simply prompting an off-the-shelf LLM to generate a zero-shot task description already achieves a strong baseline performance. However, for \emph{unknown or dynamic tasks} where the specific objective cannot be predefined, we introduce an evolutionary prompt optimization method. By analyzing a small calibration set of queries, this method automatically infers the underlying user intention and iteratively refines the task guide. 
% Consequently, our framework empowers users to generate effective, task guide on the fly removing the burden of manual task definition.

Furthermore, we repurpose scale factors from quantization~\cite{liu2024kivi} as a zero-cost proxy to identify influential outlier tokens within Value. Incorporating this outlier detection, TaskPress achieves highly efficient compression, which
we empirically validate on LongBench \cite{bai2024longbench} and RULER \cite{hsieh2024ruler} over both query-agnostic and query-dependent baselines.

\section{Related Work}
\label{sec:related_work}

\paragraph{KV Cache Pruning and Eviction.} 
%The linear expansion of the KV cache is a fundamental bottleneck for scaling Large Language Models (LLMs) to long contexts. Early strategies to constrain cache size relied on 
Early approaches rely on static or heuristic-based token eviction:
%For instance, 
StreamingLLM \cite{xiao2023efficient} retains initial attention sinks to stabilize generation for infinite-length inputs, while H$_2$O \cite{zhang2023h2o} drops tokens with low accumulated attention scores. Building on these foundations, more recent work shifts toward \textit{query-aware} pruning. Approaches such as SnapKV \cite{li2024snapkv}, PyramidKV \cite{cai2024pyramidkv}, and Quest \cite{tang2024quest} compute token importance dynamically based on the attention allocation from a specific user query. Although these query-dependent methods achieve high compression rates for single-turn inference,
%the resulting cache is strictly overfitted to the observed query. Consequently, the 
compressed memory cannot be safely reused for new queries over the same context, enforcing redundant and expensive prefill computations in multi-turn or multi-query workloads.

\paragraph{Query-Agnostic  Compression.}
To overcome the limitations of query-dependent pruning, query-agnostic KV cache management approaches include Expected Attention \cite{devoto2025expected},  estimating token importance based on the anticipated distribution of future queries. Alternatively, KVZip \cite{kim2025kvzip} addresses cache reusability by introducing a context reconstruction to adapt compressed cache to novel queries. However, such reconstruction doubles the prompt length during the prefill stage, incurring a severe quadratic computational overhead $\sim\mathcal{O}(L_c^2)$. 
%TaskPress introduces a paradigm shift by leveraging a high-level \textit{task guide} as a structural meta-query. This decoupling ensures query-agnosticism and high reusability while maintaining a minimal, linear-time complexity $\mathcal{O}(L_t L_c)$ during prefill.

\begin{comment}
\paragraph{Our Distinction.}
Quantization techniques, e.g., KIVI \cite{liu2024kivi}, SpinQuant \cite{liu2024spinquant} reduce memory bandwidth pressure and preserve outliers in activations. However, they ignore Value states entirely or rely on Value-centric metrics \cite{sengupta2025value} that incur significant latency. 
%TaskPress uniquely resolves this by repurposing pre-computed quantization scale factors---already available in standard serving pipelines---as a zero-cost proxy to detect high-entropy Value outliers.
Since optimizing discrete prompts via continuous gradients is intractable, methods like GEPA \cite{agrawal2025gepa} utilize reflective evolutionary algorithms. TaskPress adopts this paradigm to iteratively update a high-level task guide. Through this evolutionary refinement.
%TaskPress progressively maximizes the mutual information between the task guide and anticipated future queries. Consequently, the evolved guide acts as an optimal \textit{information surrogate}, explicitly framing query-agnostic cache compression as a task-time memory optimization problem.

\end{comment}
\paragraph{Our Distinction.}
Ours is a query-agnostic approach with distinction of using task guide as a comprehensive query encapsulating anticipated future queries.
% maximizing the mutual information between the task guide and anticipated future queries.
%While quantization techniques \cite{liu2024kivi,liu2024spinquant}) are typically employed to reduce memory bandwidth pressure and preserve activation outliers,
Another distinction of TaskPress is repurposing pre-computed quantization scale factors~ \cite{liu2024kivi,liu2024spinquant} readily available in serving pipelines, as a zero-cost proxy to detect high-entropy Value outliers.
Lastly, TaskPress employs an evolutionary search to discover the optimal task guide for unknown task inspired by prompt optimizer like GEPA \cite{agrawal2025gepa}.

\section{Method}
\label{sec:method}

% We introduce \textbf{TaskPress} with following distinctions:
% %a framework designed to compress the KV cache by retaining only the tokens most relevant to the underlying task and those containing critical outlier features.
% \begin{itemize}
% \item key: guided by meta-query estimating expected query distribution
% %downstream queries estimates 
% \item value: quantization scale
% \item TaskPress combines these key and value importance signals and searches for optimal task guide 
% %to compute a task-conditioned
% %token score used for KV cache pruning. Because the effectiveness of pruning depends on
% %the choice of task guide, we further introduce an evolutionary optimization procedure
% %that automatically refines guides based on downstream task performance.
% \end{itemize}
% We introduce \textbf{TaskPress} with the following  core mechanisms:

% \begin{itemize}
%     \item \textit{key} importance guided by a meta-query that estimates the expected distribution of downstream queries (Section~\ref{sec:task_attn}) and \textit{value} importance  by the quantization scale  (Section~\ref{sec:value_scale}) 
%     \item Multiplicative scoring using key and value importances (Section~\ref{sec:unified_score}) and task guide discovery using evolutionary search   (Section~\ref{sec:evolution}).
% \end{itemize}

We introduce \textbf{TaskPress}, a framework driven by four core mechanisms. First, we establish \textit{key} importance guided by a meta-query that estimates the expected distribution of downstream queries (Section~\ref{sec:task_attn}). Second, we determine \textit{value} importance derived from the quantization scale (Section~\ref{sec:value_scale}). Third, we integrate these metrics through a multiplicative scoring function (Section~\ref{sec:unified_score}). Finally, we employ an evolutionary search algorithm for task guide discovery (Section~\ref{sec:evolution}).

\subsection{Problem Statement}
\label{sec:prel}

%Unlike reactive methods \cite{li2024snapkv} that recalculate token importance for each new query, we utilize the \textit{task guide} as a stable \textbf{``meta-query.''} Since all 

Utilizing a task guide $T$ as a stable meta-query,
we decouple compression from specific query, enabling the creation of a query-agnostic subset that can be reused by 
future user queries aligned with the task's objective
 (We detail the evolutionary optimization of this optimal guide $T$ in Section \ref{sec:evolution}.) 

Given a KV cache of length $L_c$.
our goal is to select a subset of indices of context $\mathcal{I} \subset \{1, \dots, L_c\}$ for each head with cardinality $|\mathcal{I}| \ll L_c$, such that the compressed KV cache preserves performance on subsequent queries governed by its task.

\subsection{Task-Guided Key Importance}
\label{sec:task_attn}

%Unlike reactive methods \cite{li2024snapkv} that recalculate token importance for each new query, we utilize a task guide $T$ as a stable meta-query. (We detail the evolutionary optimization of this optimal guide $T$ in Section \ref{sec:evolution}.) 

During the prefill phase, the task description of length $L_t$ is appended to the context of length $L_c$; subsequently, the KV cache corresponding to the task description is evicted immediately after prefill.

Let $\mathbf{A} \in \mathbb{R}^{L_t \times L_c}$ denote the attention map between the task description and input context, sliced from the entire attention map of size $L_{c+t}\times L_{c+t}$. $\mathbf{A}_{j,i}$ represents the weight assigned to context token $i$ by task token $j$. By aggregating scores across the task dimension and applying 1D average pooling with kernel size $K$ to capture local dependencies, the task-guided score for the $i$-th context token is:

\begin{equation}
s^{\text{key}}_i = \frac{1}{K}\sum_{k=-\lfloor K/2 \rfloor}^{\lfloor K/2 \rfloor} \sum_{j=1}^{L_t} \mathbf{A}_{j, i+k}
\end{equation}

Since this derives strictly from query-key interactions, $s^{\text{key}}_i$ exclusively measures Key importance.

\subsection{Quantization Scale as Value Importance}
\label{sec:value_scale}

However, the task-guided Key score does not measure the density of the retrieved value.
Since self-attention operates as a weighted sum of attention weights and value, capturing the importance of value is vital as well.
To address this, we propose to identify activation outliers, which act as effective indicators of high-entropy, critical features. 

Our distinction is reusing quantization scale factors as such measure. Specifically, assuming a per-token quantization \cite{liu2024kivi} scheme, the scale factor $\gamma_i$ of a value $\mathbf{v}_i$ is calculated as:
\begin{equation}
\gamma_i = \frac{\max(|\mathbf{v}_i|)}{2^{b-1} - 1}
\end{equation}

\noindent Since the bit-width $b$ is constant across all tokens, $\gamma_i$ is directly proportional to the maximum magnitude of $\mathbf{v}_i$. We therefore define our value outlier score as $s^{\text{value}}_i = \gamma_i$. A key advantage of this approach is its zero-overhead: because quantization is the de facto standard for efficient LLM serving, $\gamma_i$ is already computed and readily available in memory, requiring no additional  operations. We've found that $\gamma_i$ is an efficient proxy for the norm and discussed further on Section \ref{sec:value_score_justification}.

\subsection{KV Importance Scoring}
\label{sec:unified_score}
To construct the final task-conditioned memory, the importance score for cache retention is computed as the element-wise product of the task-guided key importance and the value outlier score: 
\begin{equation}
s_i = s^{\text{key}}_i \cdot s^{\text{value}}_i
\end{equation}

By employing a multiplicative interaction, it must concurrently satisfy two conditions: it must be \textit{structurally retrievable} via the task-aligned attention space (high Key score) and \textit{intrinsically rich} in semantic information (high Value outlier score). Finally, the indices $\mathcal{I}$ corresponding to the tokens with the highest combined scores $s_i$ are selected.

\begin{table*}[t]
\centering
\caption{LongBench results with zero-shot task guide. Eviction Ratio denotes the percentage of the KV cache that has been pruned. Boldface and underline denote the best and the second-best accuracy, respectively. $\cdot^{\texttt{int4}}$ denotes quantized KV cache. By default, all methods use quantized KV cache.}
\vspace{-2mm}
\resizebox{\textwidth}{!}{%
\begin{tabular}{llc ccc ccc ccc ccc cc c}
\toprule
\multirow{2}{*}{} & & \multirow{2}{*}{\shortstack{Eviction\\Ratio}} & \multicolumn{3}{c}{Single-Document QA} & \multicolumn{3}{c}{Multi-Document QA} & \multicolumn{3}{c}{Summarization} & \multicolumn{3}{c}{Few-shot Learning} & \multicolumn{2}{c}{Synthetic} & \multirow{2}{*}{Average} \\
\cmidrule(lr){4-6} \cmidrule(lr){7-9} \cmidrule(lr){10-12} \cmidrule(lr){13-15} \cmidrule(lr){16-17}
 & & & NrtvQA & Qasper & MF-en & HotpotQA & 2WikiMQA & Musique & GovReport & QMSum & MultiNews & TREC & TriviaQA & SAMSum & PCount & PRe & \\
\midrule
\multirow{15}{*}{\rotatebox{90}{LLaMA-3.1-8B-Instruct}} 
 & All KV & 0\% & 30.72 & 47.06 & 55.28 & 59.51 & 51.83 & 32.66 & 35.25 & 24.94 & 27.05 & 29.5 & 85.81 & 38.88 & 10.7 & 100.0 & 44.94 \\
 & All KV$^{\texttt{int4}}$ & 0\% & 30.54 & 47.11 & 55.64 & 58.6 & 48.34 & 33.61 & 34.99 & 24.82 & 26.75 & 27.0 & 82.91 & 39.9 & 11.5 & 99.5 & 44.37 \\
\cmidrule(lr){2-18}
 & SnapKV & \cellcolor{blue!0} 50\% & 29.28 & 40.64 & 46.36 & \textbf{56.9} & \textbf{48.96} & \underline{28.85} & 31.87 & 23.8 & 25.64 & 36.5 & 82.46 & \textbf{41.13} & 11.0 & \textbf{99.0} & 43.03 \\
 & PyramidKV & \cellcolor{blue!0} 50\%  & 28.48 & 36.34 & 45.08 & 48.48 & 38.81 & 23.32 & 31.0 & 22.57 & 25.37 & 49.5 & 84.43 & 32.1 & 9.18 & 96.92 & 40.83 \\
 & StreamingLLM & \cellcolor{blue!0} 50\%  & 24.92 & 39.25 & 31.78 & 48.99 & 38.72 & 24.88 & 30.68 & 22.42 & 25.77 & 35.0 & \textbf{91.7} & 37.6 & 8.88 & 54.0 & 36.76 \\
 & KVzip & \cellcolor{blue!0} 50\%  & \textbf{32.15} & \textbf{44.8} & \textbf{55.65} & 53.28 & 48.08 & 25.1 & \underline{33.25} & \textbf{24.56} & \textbf{26.61} & \underline{53.0} & \underline{86.74} & 38.97 & \underline{11.39} & 93.0 & \underline{44.76} \\
 & TaskPress (Ours) & \cellcolor{blue!0} 50\%  & \underline{30.65} & \underline{42.12} & \underline{54.78} & \underline{56.55} & \underline{48.32} & \textbf{34.62} & \textbf{33.68} & \underline{23.97} & \underline{25.95} & \textbf{63.5} & 82.81 & \underline{40.73} & \textbf{13.67} & \underline{98.5} & \textbf{46.42} \\
\cmidrule(lr){2-18}

 & SnapKV & \cellcolor{blue!0} 75\% & 28.76 & 30.27 & 35.96 & 55.17 & \textbf{43.56} & 26.34 & 29.08 & 22.06 & 23.54 & 33.5 & 82.48 & \textbf{41.34} & 8.1 & 91.0 & 39.37 \\
 & PyramidKV & \cellcolor{blue!0} 75\%  & 23.06 & 28.96 & 35.04 & 40.72 & 36.32 & 18.34 & 27.93 & 20.96 & 23.42 & 41.0 & 87.84 & 33.41 & 9.71 & 80.92 & 36.26 \\
 & StreamingLLM & \cellcolor{blue!0} 75\%  & 24.05 & 24.24 & 25.22 & 43.34 & 27.16 & 21.14 & 28.88 & 20.1 & 23.45 & 34.0 & \textbf{91.03} & 36.21 & 6.5 & 33.5 & 31.34 \\
 & Expected Attention & \cellcolor{blue!0} 75\% & \underline{29.22} & \underline{40.36} & 42.93 & \underline{55.45} & \underline{43.45} & 28.17 & \textbf{31.71} & 23.94 & \textbf{26.2} & 29.5 & 87.68 & 38.11 & \underline{10.33} & 48.0 & 38.22 \\
 & ThinkPress & \cellcolor{blue!0} 75\% & 5.99 & 8.06 & 19.15 & 22.1 & 11.74 & 6.64 & 19.45 & 17.17 & 16.67 & 7.5 & 88.15 & 31.43 & 0 & 4.5 & 18.47 \\
 & KVzip & \cellcolor{blue!0} 75\% & 28.71 & 35.83 & \underline{48.54} & 50.42 & 39.86 & 22.59 & 30.66 & \textbf{24.73} & \underline{25.66} & 20.0 & 87.84 & 37.89 & 9.07 & 73.5 & 38.24 \\
 & TaskPress (Ours) & \cellcolor{blue!0} 75\% & \textbf{30.37} & 32.29 & 45.76 & 55.13 & 42.04 & \underline{29.77} & \underline{31.28} & 23.61 & 24.11 & \textbf{61.5} & 77.48 & \underline{41.26} & \textbf{11.0} & \underline{97.5} & \underline{43.08} \\
 & TaskPress + ThinkPress (Ours) & \cellcolor{blue!0} 75\% & 27.81 & \textbf{40.42} & \textbf{51.03} & \textbf{58.28} & 40.46 & \textbf{32.03} & \textbf{31.71} & \underline{24.4} & 25.14 & \underline{59.5} & \underline{89.7} & 36.27 & \textbf{11.0} & \textbf{98.0} & \textbf{44.70} \\
\midrule

\multirow{15}{*}{\rotatebox{90}{Qwen3-8B}} 
 & All KV & 0\% & 28.77 & 42.04 & 53.62 & 62.29 & 47.17 & 34.16 & 33.65 & 24.31 & 24.76 & 46.5 & 88.49 & 40.67 & 10.19 & 90.98 & 44.83 \\
 & All KV$^{\texttt{int4}}$ & 0\% & 28.83 & 44.8 & 55.6 & 62.94 & 49.16 & 35.44 & 33.59 & 24.53 & 24.69 & 41.0 & 90.31 & 40.34 & 10.0 & 91.43 & 45.19 \\
\cmidrule(lr){2-18}

 & SnapKV & \cellcolor{green!0} 50\% & \underline{26.91} & \underline{36.01} & 44.83 & \textbf{59.46} & \textbf{46.0} & \underline{32.34} & 32.28 & 22.6 & 23.4 & 44.5 & \textbf{88.59} & \underline{40.52} & 8.99 & 92.47 & 42.78 \\
 & PyramidKV & \cellcolor{green!0} 50\%  & 25.55 & 29.77 & 37.59 & 55.81 & 37.39 & 31.24 & 30.0 & 21.03 & 21.57 & 54.0 & 87.81 & 38.27 & 9.5 & \textbf{98.83} & 41.31 \\
 & StreamingLLM & \cellcolor{green!0} 50\%  & 23.24 & 31.72 & 30.89 & 47.56 & 37.03 & 20.68 & 31.12 & 21.6 & \underline{24.05} & 47.5 & 72.99 & 38.53 & \underline{10.5} & 53.92 & 35.10 \\
 & KVzip & \cellcolor{green!0} 50\%  & \underline{26.36} & \textbf{42.72} & \underline{49.27} & 56.77 & \textbf{46.0} & 29.42 & \underline{32.76} & \textbf{23.39} & \textbf{24.58} & \underline{67.5} & 87.35 & 39.63 & 10.13 & 83.02 & \underline{44.21} \\
 & TaskPress (Ours) & \cellcolor{green!0} 50\%  & \textbf{27.43} & 35.67 & \textbf{49.49} & \underline{58.59} & 43.15 & \textbf{36.44} & \textbf{33.06} & \underline{23.31} & 23.43 & \textbf{68.5} & \underline{88.02} & \textbf{40.61} & \textbf{11.39} & \underline{93.06} & \textbf{45.15} \\
\cmidrule(lr){2-18}
 & SnapKV & \cellcolor{green!0} 75\% & 24.77 & 28.44 & 35.09 & 50.71 & 38.33 & 26.86 & 29.34 & 21.06 & 21.38 & 41.25 & \underline{88.49} & \underline{40.39} & 9.83 & 87.52 & 38.82 \\
 & PyramidKV & \cellcolor{green!0} 75\%  & 22.39 & 26.78 & 32.47 & 48.88 & 33.34 & 23.78 & 27.55 & 19.76 & 21.31 & 42.25 & 88.24 & 38.06 & 8.0 & 85.42 & 37.02 \\
 & StreamingLLM & \cellcolor{green!0} 75\%  & 21.8 & 20.72 & 27.6 & 41.46 & 30.94 & 19.41 & 28.74 & 20.04 & 21.4 & 29.5 & 72.55 & 33.62 & 9.0 & 35.5 & 29.45 \\
 & Expected Attention & \cellcolor{green!0} 75\% & \textbf{27.28} & 34.55 & 39.74 & \underline{57.79} & 39.19 & 28.03 & \underline{32.2} & 22.89 & \underline{23.89} & 66.0 & 86.14 & 39.31 & 10.07 & 48.33 & 39.67 \\
 & ThinkPress & \cellcolor{green!0} 75\% & 18.84	&15.97&	30.88&	31.22&	22.86	&9.41	&21.59&	18.43&	17.13&	48.5	&86.56	&32.95&	7.0 &	14.67	&26.86 \\
 & KVzip & \cellcolor{green!0} 75\% & 25.07 & \textbf{38.15} & \underline{44.14} & 50.87 & 36.97 & 26.17 & 31.49 & \underline{23.0} & \textbf{24.19} & 46.17 & 87.24 & 38.47 & \textbf{11.0} & 49.12 & 38.00 \\
 & TaskPress (Ours) & \cellcolor{green!0} 75\% & 24.95 & 30.97 & 40.98 & 55.08 & \underline{39.42} & \underline{31.85} & 31.99 & 21.94 & 21.8 & \textbf{68.0} & 87.21 & \textbf{40.49} & \underline{10.7} & \underline{92.07} & \underline{42.68} \\
 & ThinkPress + TaskPress (Ours) & \cellcolor{green!0} 75\% & \underline{26.48} & \underline{36.27} & \textbf{48.29} & \textbf{59.23} & \textbf{42.1} & \textbf{35.8} & \textbf{33.09} & \textbf{23.19} & 22.78 & \underline{67.0} & \textbf{88.74} & 39.8 & 10.5 & \textbf{96.14} & \textbf{44.96} \\
\bottomrule
\end{tabular}%
}
\label{tab:longbench_results}
\end{table*}

\subsection{Task Guide Formulation}~\label{ref:gen}
In this section, we propose a systematic, end-to-end framework to construct an optimal task guide $T^*$. 
A task guide encapsulates where future queries will focus. Therefore, the task guide should define semantic scope or entity to focus on, such as character interactions, dates, or names.

\subsubsection{Zero-shot task guide initialization}
\label{method:task_init}
We utilize existing conversational LLMs to generate a strong, knowledge-grounded baseline guide for initialization. We leverage an off-the-shelf Gemini app to generate a concise summary of the task objective. Specifically, for each task we used this prompt template: \texttt{``Generate a task guidance for \{task\} in 2--3 sentences.''}. 
The resulting natural language is used directly as the task guide for the subsequent attention computation with minimal modification.

% Delta_Begin
% In our experiment, the initial task guides generated by off-the-shelf LLMs already serve as a strong static baseline (Section \ref{experiment:longbench}). However, a zero-shot prompt inherently struggles to capture the diverse, latent semantics of actual user queries. Our objective is to discover the optimal task guide $T^*$ that maximizes mutual information with those queries. 

 In our experiment, the initial task guides generated by off-the-shelf LLMs already serve as a strong static baseline (Section \ref{experiment:longbench}). 
 
 % However, initial zero-shot task guide might be drifted from actual user queries and also LLM won't be able to construct task guide for new or unknown tasks.
 % Therefore, we propose a evolutionary task guide discovery to discover the optimal task guide $T^*$ that maximizes mutual information with those queries. 

\subsubsection{Evolutionary task guide discovery} 
\label{sec:evolution}

% For unknown tasks where a predefined objective is unavailable to LLM, rendering the zero-shot task is inapplicable. For such scenarios, we propose to dynamically generate the task guide by inferring the user intent directly from a small collection of proxy queries. By analyzing these representative queries, our goal is to extract the common informational needs and construct a unifying meta-query.

% Because directly optimizing natural language through gradient descent is intractable, we circumvent this limitation by employing an evolutionary Algorithm \ref{alg:evolve}. This approach systematically refines the guide prior to inference using a small calibration set $\mathcal{D}_{\text{cal}}$. 
% The evolutionary process iteratively optimizes a population of candidate task guides. In each generation, candidates are evaluated and top-performing guides selected based on downstream task accuracy on $\mathcal{D}_{\text{cal}}$. 
%  By repeating this cycle, the algorithm discovers a optimized meta-query $T^*$.

% Furthermore, this approach can be utilized to refine the task guide even when the task is already known. By initializing $\mathcal{T}_0$ with the zero-shot task guide, we demonstrate the improvements achieved through this refinement in Table~\ref{tab:evolution_results}.

For unknown tasks, zero-shot prompts are inapplicable because LLM does not aware of task's objective. To address this, we propose generating a task guide by inferring intent from a small collection of queries, consolidating their shared informational needs into a single unifying meta-query.

Because directly optimizing natural language via gradient descent is intractable, we adopt an evolutionary Algorithm \ref{alg:evolve} to iteratively refine candidate guides. Operating on a small calibration set ($\mathcal{D}_{\text{cal}}$), the algorithm evaluates each generation, selects top-performing candidates based on downstream task accuracy, and converges on an optimized meta-query $T^*$. 
Furthermore, this framework can be repurposed for zero-shot task guide refinement. Our evolutionary refinement yields  accuracy gains, as demonstrated in Table~\ref{tab:evolution_results}.

% \begin{algorithm}[t]
% \caption{Task Guide Evolution}
% \begin{algorithmic}[1]

% \REQUIRE Initial guide $T_\text{base}$, Calibration set $\mathcal{D}_\text{cal}$

% \REQUIRE LLM mutator $\mathcal{M}$, Total generations $G$

% \REQUIRE Selection size $K$, Number of new candidates $N$

% \STATE \textbf{Initialize Candidate Pool $\mathcal{C}$:} 
% \STATE \# Sample calibration queries $Q_\text{cal} \sim \mathcal{D}_\text{cal}$
% \STATE \# Generate $N$ initial candidates:
% \STATE $\mathcal{C}_0 \leftarrow \mathcal{M}(\{T_\text{base}\}, Q_\text{cal}; N) \cup \{T_\text{base}\}$

% \FOR{$g = 1$ \TO $G$}

%     \STATE \textbf{Evaluate \& Select:} 
%     \STATE $\mathcal{C}_{\text{best}} \leftarrow \operatorname{Top-}K(\mathcal{C}_{g-1}, \text{Score}(\cdot, \mathcal{D}_\text{cal}))$

%     \STATE \textbf{Mutate:}

%     \STATE \# Add $N$ new candidates:
%     \STATE $\mathcal{C}_g \leftarrow \mathcal{M}(\mathcal{C}_{\text{best}}, Q_\text{cal}; N)  \cup \mathcal{C}_{\text{best}} $

% \ENDFOR

% \STATE \# Find the best task guide:

% \RETURN $T^* = \arg\max_{T \in \mathcal{C}_G} \text{Score}(T, \mathcal{D}_\text{cal})$

% \end{algorithmic}

% \label{alg:evolve}

% \end{algorithm}

\vspace{0cm}
\begin{algorithm}[htpb]
\caption{Task Guide Discovery}
\begin{algorithmic}[1]

\REQUIRE Calibration set $\mathcal{D}_\text{cal}$, LLM mutator $\mathcal{M}$, Total generations $G$, Selection size $K$
\STATE \textbf{Initialize Candidate Pool $\mathcal{T}$:} 
\STATE \# Sample calibration queries $Q_\text{cal} \sim \mathcal{D}_\text{cal}$
\STATE \# Generate $N$ initial candidates:
\STATE $\mathcal{T}_0 \leftarrow \mathcal{M}( Q_\text{cal}; N)$

\FOR{$g = 1$ \TO $G$}

    \STATE \textbf{Evaluate \& Select:} 
    \STATE $\mathcal{T}_{\text{best}} \leftarrow \operatorname{Top-}K(\mathcal{T}_{g-1}, \text{Score}(\cdot, \mathcal{D}_\text{cal}))$

    \STATE \textbf{Mutate}

    \STATE \# Add $N$ new candidates:
    \STATE $\mathcal{T}_g \leftarrow \mathcal{M}(\mathcal{C}_{\text{best}}, Q_\text{cal}; N)  \cup \mathcal{T}_{\text{best}} $

\ENDFOR

\STATE \# Return the best task guide:

\RETURN $T^* = \arg\max_{T \in \mathcal{T}_G} \text{Score}(T, \mathcal{D}_\text{cal})$

\end{algorithmic}

\label{alg:evolve}

\end{algorithm}

\vspace{-0.5cm}

\section{Experiments}

% \subsection{Setup}
\label{exp:setup}

\noindent\textbf{Baselines.} We compare ours against recent state-of-the-art methods SnapKV \cite{li2024snapkv}, PyramidKV \cite{cai2024pyramidkv}, StreamingLLM \cite{xiao2023efficient}, ExpectedAttention \cite{devoto2025expected}, ThinK \cite{xu2025think} and KVzip \cite{kim2025kvzip}  using LLaMA \cite{dubey2024llama} and Qwen3 \cite{yang2025qwen3} models.

\noindent\textbf{Quantization.} To simulate realistic deployment, we apply \texttt{int4} KV cache quantization \cite{liu2024spinquant}. There were negligible impacts on quality.
This yields 4$\times$ compression by default; combined with 75\% eviction total compression reaches 16$\times$.

\noindent\textbf{Eviction.} We assign uniform budget for each KV head and indices $\mathcal{I}$ are selected per head independently. 
Eviction was performed after the prefill  optimizing the decoding process. The task guide prompt is utilized exclusively during the eviction phase and is discarded immediately thereafter.

\noindent\textbf{Evolution.} We retain the top $K=2$  and generate new $N=3$ candidates. The evolutionary process is conducted for $G=4$ generations.

\noindent\textbf{Code.} We conducted all of our experiments on top of 
kvpress github repository \cite{devoto2025expected} with official implementations of other methods.

\subsection{Evaluation on Long-Context Benchmarks}
\subsubsection{Results of zero-shot task guide}
\label{experiment:longbench}

We assess our method's query-agnostic long-context capabilities using the LongBench. 
For this specific evaluation, TaskPress employs only the initial, zero-shot task guides generated directly by an off-the-shelf LLM to demonstrate that task guide serves as a strong baseline even without optimizations. Following established conventions \cite{kim2025kvzip,sengupta2025value}, baselines such as SnapKV and PyramidKV utilize the last context window instead of an actual query.

As shown in Tables \ref{tab:longbench_results} and \ref{tab:longbench_large}, TaskPress outperforms baselines in most scenarios, achieving the highest overall average. Under aggressive 75\% compression, standalone channel-wise pruning like ThinK \cite{xu2025think} suffers a severe performance drop. However, combining TaskPress with ThinK (each evicting 50\%, resulting in 75\% total compression) yields the best performance. This demonstrates that a simple zero-shot LLM guide serves as a highly effective ``meta-query''.
\begin{table*}[htbp]
\centering
\caption{LongBench results of a large model Qwen3-32B on Single and Multi-Document QA benchmarks.}
\label{tab:longbench_large}
% --- Compacting commands ---
\small 
\setlength{\tabcolsep}{4pt} 
\renewcommand{\arraystretch}{0.9} 
% ---------------------------

\begin{tabular}{l ccc ccc c}
\toprule
\textbf{Qwen3-32B} & \multicolumn{3}{c}{\textbf{Single-Document QA}} & \multicolumn{3}{c}{\textbf{Multi-Document QA}} & \textbf{Average} \\
\cmidrule(lr){2-4} \cmidrule(lr){5-7}
 & NrtvQA & Qasper & MF-en & HotpotQA & 2WikiMQA & Musique & \\
\midrule
No Eviction                   & 32.99 & 47.17 & 51.86 & 53.25 & 53.70 & 29.22 & 44.70 \\
SnapKV                        & 25.88 & 33.75 & 37.44 & 51.18 & 41.69 & 25.75 & 35.95 \\
KVzip                         & 28.80 & 36.17 & 40.63 & 45.67 & 41.26 & 21.91 & 35.74 \\
TaskPress                     & 29.12 & 36.65 & 39.75 & 56.44 & 46.18 & 26.89 & 39.17 \\
TaskPress + ThinK & 28.49 & 42.64 & 49.10 & 54.84 & 51.11 & 29.24 & 42.57 \\
\bottomrule
\end{tabular}
\end{table*}

% \subsubsection{Results of evolved task guide}

% We further improved the zero-shot task guide through our proposed evolutionary optimization flow. Since LongBench lacks a dedicated validation split, we utilized 50 samples from test set as a calibration set for the optimization and we excluded these calibration samples from the final test set to ensure a fair evaluation. We keep eviction rate 50\% for this experiments.
% As shown in Table~\ref{tab:evolution_results}, our experiments on LongBench QA tasks demonstrate that the evolved task guide consistently outperforms the baseline.

\subsubsection{Results of evolved task guide}

% \begin{table*}[t]
% \centering
% \small
% \caption{Impact of evolutionary task guide optimization on QA tasks. \textbf{TaskPress (Evolved)} indicates the performance after task guide evolution. Values in parentheses highlight the absolute score improvement over the baseline TaskPress, demonstrating that optimizing the task guide provides substantial reasoning capability gains.}
% \begin{tabular}{llcccc}
% \toprule
% \textbf{Model} & \textbf{Dataset} & \textbf{StreamingLLM} & \textbf{SnapKV} & \textbf{TaskPress (Known)} & \textbf{TaskPress (Evolved)} \\
% \midrule
% \multirow{3}{*}{\shortstack[l]{Qwen3\\8B}} 
% & HotpotQA & 47.41 & 60.04 & 58.62 & \textbf{60.50} {\footnotesize (+1.88)} \\
% & MultiFieldQA & 32.38 & 46.75 & 49.44 & \textbf{50.66} {\footnotesize (+1.22)} \\
% & Qasper & 32.96 & \textbf{36.73} & 35.08 & 34.63 {\footnotesize (-0.45)} \\
% \midrule
% \multirow{3}{*}{\shortstack[l]{Llama-3.1\\8B-Instruct}} 
% & HotpotQA & 49.94 & 57.45 & 58.30 & \textbf{58.44} {\footnotesize (+0.14)} \\
% & MultiFieldQA & 35.32 & 48.90 & 51.07 & \textbf{51.32} {\footnotesize (+0.25)} \\
% & Qasper & 39.66 & 42.12 & 42.87 & \textbf{45.95} {\footnotesize (+3.08)} \\
% \bottomrule
% \end{tabular}
% \vspace{2mm}
% \label{tab:evolution_results}
% \vspace{-5mm}
% \end{table*}

% Remember to include \usepackage{arydshln} in your preamble for the dotted lines (:)

\begin{table*}[t]
\centering
\small
% \caption{Impact of evolutionary task guide optimization on QA tasks. \textbf{${\cdot}^{*}$} indicates the performance after task guide evolution. Unknown means the task guide is constructed solely from query sets, and known means task guide is initialized with zero-shot task guide prompt
% }
% \caption{Impact of evolutionary task guide optimization on QA tasks. An asterisk ($^*$) denotes performance after task guide evolution. \textbf{Unknown} indicates a scenario where TaskPress is deployed on and the guide is discovered solely from query sets. \textbf{Known} indicates task guide initialization using zero-shot prompts following the Section \ref{method:task_init}.  Values in parentheses highlight the absolute score improvement over (Known)  task guide.}
\caption{
Impact of evolutionary task guide optimization on QA tasks. $^*$ refers to an experiment with an evolved task guide. \textbf{Unknown} refers to task guides discovered solely from query sets. \textbf{Known} refers to task guides with zero-shot initialization (Sec. \ref{method:task_init}). Values in parentheses highlight the absolute score improvement from evolution.}
\begin{tabular}{llcc|c|cc}
\toprule
% \textbf{Model} & \textbf{Dataset} & \textbf{StreamingLLM} & \multicolumn{1}{c}{\textbf{SnapKV}} & \multicolumn{1}{c}{\shortstack[c]{\textbf{TaskPress}^{*} \\ \textbf{(Unknown)}}} & \shortstack[c]{\textbf{TaskPress} \\ \textbf{(Known)}} & \shortstack[c]{\textbf{TaskPress}^{*} \\ \textbf{(Known)}} \\
\textbf{Model} & \textbf{Dataset} & \textbf{StreamingLLM} & \multicolumn{1}{c}{\textbf{SnapKV}} & \multicolumn{1}{c}{\shortstack[c]{\textbf{TaskPress}$^{*}$ \\ \textbf{(Unknown)}}} & \shortstack[c]{\textbf{TaskPress} \\ \textbf{(Known)}} & \shortstack[c]{\textbf{TaskPress}$^{*}$ \\ \textbf{(Known)}} \\
\midrule
\multirow{3}{*}{\shortstack[l]{Qwen3\\8B}} 
& HotpotQA & 47.41 & 60.04 & 59.08 & 58.62 & \textbf{60.50} {\footnotesize (+1.88)} \\
& MultiFieldQA & 32.38 & 46.75 & 49.00 & 49.44 & \textbf{50.66} {\footnotesize (+1.22)} \\
& Qasper & 32.96 & \textbf{36.73} & 36.50 & 35.08 & 34.63 {\footnotesize (-0.45)} \\
\midrule
\multirow{3}{*}{\shortstack[l]{Llama-3.1\\8B-Instruct}} 
& HotpotQA & 49.94 & 57.45 & \textbf{58.49} & 58.30 & 58.44 {\footnotesize (+0.14)} \\
& MultiFieldQA & 35.32 & 48.90 & \textbf{51.83} & 51.07 & 51.32 {\footnotesize (+0.25)} \\
& Qasper & 39.66 & 42.12 & 42.64 & 42.87 & \textbf{45.95} {\footnotesize (+3.08)} \\
\bottomrule
\end{tabular}
\vspace{2mm}
\label{tab:evolution_results}
\vspace{-5mm}
\end{table*}

% To simulate a scenario where TaskPress is deployed on unknown tasks, we evaluate the proposed evolutionary discovery flow. Instead of explicitly instructing the LLM to generate a guide for a predefined task, we allow the model to infer the underlying objective directly from a set of queries. Because LongBench lacks a dedicated validation split, we repurpose 50 samples from the test set to serve as a calibration set for optimization. To ensure a fair evaluation, we strictly exclude these calibration samples from the final test set. We maintain an eviction rate of 50\% for these experiments. As shown in Table~\ref{tab:evolution_results}, our experiments on LongBench QA tasks demonstrate that our pipeline can discover effective task guide for unknown tasks.

% Moreover, we also applied the evolutionary algorithm on known task's task guide. And the results shows that the proposed method effectively improve zeros-hot task guide.

To simulate a scenario where TaskPress is deployed on unknown tasks, we employ our evolutionary discovery flow, allowing the model to infer the task guide directly from queries. Lacking a dedicated LongBench validation split, we optimize using 50 calibration samples repurposed from the test set, strictly excluding them from the final evaluation. At a 50\% eviction rate, Table~\ref{tab:evolution_results} shows this pipeline successfully discovers effective guides for unknown QA tasks.  Furthermore, applying this pipeline to the zero-shot task guides (Section~\ref{method:task_init}) effectively improves overall accuracy.

\subsubsection{Results on retrieval capability}

We evaluate our method on RULER \cite{hsieh2024ruler} needle-in-a-haystack subtasks. We formed a unified task guide by concatenating short instructions for each subtasks, thereby challenging the method to handle multiple subtasks simultaneously. 

As shown in Table~\ref{tab:ruler}, we achieve best accuracy at a 75\% eviction rate. 
For 50\% eviction rate, we achieved the second best accuracy against self-reconstruction counterpart which is more costly.
While TaskPress outperforms others on most of subtasks, but it gained relatively low accuracy on MK-2/3.
We attribute this degradation to attention dilution caused by an excessive number of target needles, which we analyze in detail in Appendix~\ref{app:mk23}.

\begin{table*}[t]
\centering
\footnotesize % 표 전체 폰트 크기를 약간 줄임 (필요시 \footnotesize 로 변경 가능)
\setlength{\tabcolsep}{4pt} % 열 사이의 기본 여백(6pt)을 4pt로 줄여서 폭을 좁힘
\caption{Results on RULER-16K with LLaMA-3.1-8B-Instruct. Boldface and underline denotes the best score among 75\% and 50\% eviction rates respectively.}
\vspace{-2mm}
\begin{tabular}{lcccccc}
\toprule
 & All & SnapKV & PyramidKV & ExpectedAttention & KVZip & TaskPress \\
 Sub-Task  & \scriptsize 100\% & \scriptsize 50\% / 75\% & \scriptsize 50\% / 75\% & \scriptsize 50\% / 75\% & \scriptsize 50\% / 75\% & \scriptsize 50\% / 75\% \\
\midrule
CWE & 88.1 & 75.7 / 20.5 & 40.2 / 5.52 & 79.3 / 61.8 & 64.3 / 23.5 & \underline{86.3} / \textbf{71.1} \\
FWE & 91.7 & 89.7 / 79.0 & 86.2 / 74.9 & 85.1 / 78.4  & \underline{89.9} / 82.5  & 88.9 / \textbf{87.0} \\
MK-1 & 99.4 & 92.4 / 49.2 & 91.0 / 48.8 & 91.0 / 35.0  & 90.4 / 66.2 & \underline{99.4} / \textbf{99.6} \\
MK-2 & 100 & 43.8 / 17.8 & 39.6 / 12.2 & 68.2 / 17.6 &  \underline{99.8} / \textbf{82.4} & 50.0 / 10.6 \\
MK-3 & 99.6 & 16.8 / 3.6  & 15.4 / 3.6 & 35.0 / 2.0  & \underline{78.4} / \textbf{5.2} & 7.2 / 1.0 \\
MQ & 99.1 & 83.9 / 34.6 &  84.8 / 31.4 & 74.5 / 24.0 & 93.7 / 76.4  & \underline{99.3} / \textbf{98.6} \\
MV & 98.9 & 79.6 / 30.3 & 84.0 / 32.0 & 71.2 / 15.3 & 91.4 / 70.0 & \underline{98.9} / \textbf{98.1} \\
S-1 & 100 & 98.0 / 91.6 & 92.2 / 87.6 & 99.8 / 99.2 & \underline{100} / \textbf{100} & \underline{100} / \textbf{100} \\
S-2 & 100 & 97.6 / 81.2  & 99.6 / 82.4& 93.4 / 48.8  &  90.0 / 62.4 & \underline{100} / \textbf{100} \\
S-3 & 100 & 14.8 / 3.4 & 9.6 / 2.6 & 18.0 /6.0 &   84.6 / 29.2 & \underline{92.2} / \textbf{44.0} \\
QA-1 & 81.6 & 51.6 / 32.4 & 42.6 / 24.8 & \underline{76.0} / 63.4 & 74.8 / \textbf{64.6} & 54.8 / 37.6 \\
QA-2 & 56.8 & 44.6 / 34.4 & 32.4 / 26.2 & \underline{52.6} / 44.4 & 52.4 / \textbf{46.6} & 50.4 / 42.2 \\
VT & 99.8 & 93.8 / 78.6  & 88.64 / 77.36 & 97.8 / 77.7 & 99.8 / \textbf{99.6} & \underline{100} / 99.4 \\
\cdashline{1-7}[2pt/1pt]
Avg & 93.5 & 67.9 / 42.8 & 62.0 / 39.2 & 72.4 / 44.1  & \underline{85.3} / 62.2 & 79.0 / \textbf{68.4} \\
\bottomrule
\end{tabular}
\label{tab:ruler}
\end{table*}

\subsubsection{Comparison with query-aware method}

\begin{figure}
    \centering
    % Replace 'combined_plot.png' with your actual file name
    \includegraphics[width=1.0\linewidth]{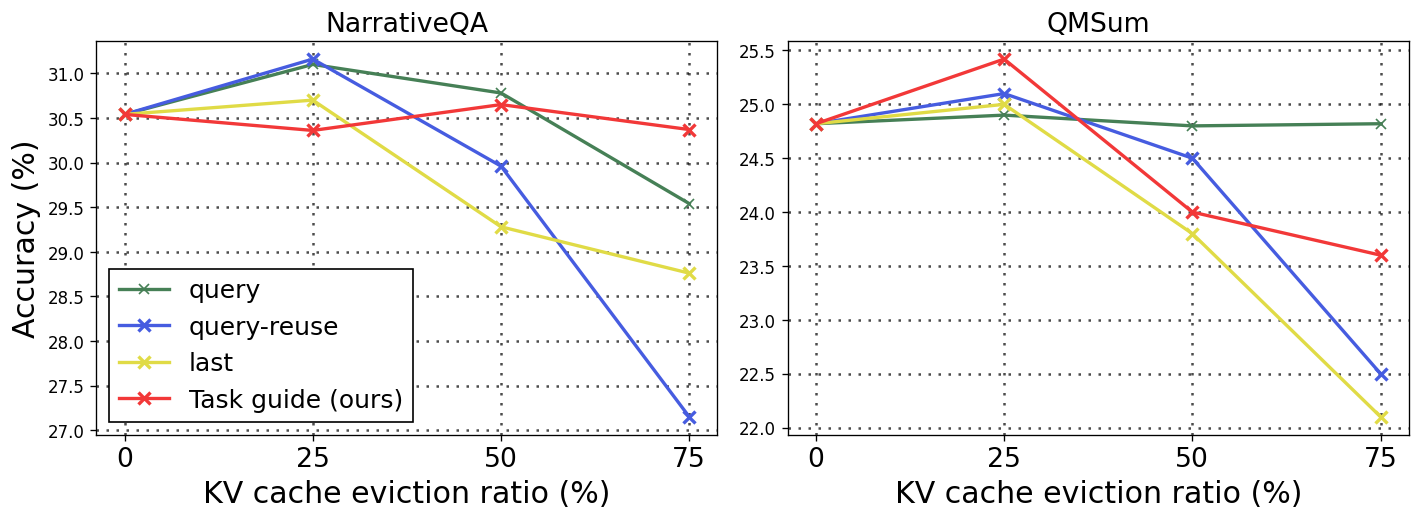}
    \caption{Accuracy comparison on the NarrativeQA and QMSum datasets with LLaMA-3.1-8B-Instruct. We evaluate four compression strategies: (1) \textit{query}-dependent, where the specific query guides eviction; (2) \textit{query-reuse}, where a cache compressed for one query is applied to others; (3) \textit{last} context, where the final context segment serves as a proxy for query; and (4) Task guide (Ours), which utilizes a high-level task guide.}
    \label{fig:query_dependent}
\end{figure}

We investigate whether task-guided KV cache eviction can still compete with query-dependent baselines in their native settings. We evaluate this on two QA datasets: NarrativeQA \cite{kovcisky2018narrativeqa} and QMSum \cite{zhong2021qmsum}. 
These benchmarks were selected because they pair a single document with multiple questions, a setting where query-dependent methods typically excel. In Fig~\ref{fig:query_dependent}, the green line labeled `query' represents the native performance of query-dependent baselines; SnapKV. Our results demonstrate that TaskPress (red line) achieves performance comparable to or better than these baselines. Specifically, even at a highly aggressive eviction rate of 75\%. 

\subsection{Impact on End-to-End Efficiency}
\subsubsection{Cost of KV cache eviction}

\begin{figure}[h]
    \centering
    % Replace 'combined_plot.png' with your actual file name
    \includegraphics[width=1.0\linewidth]{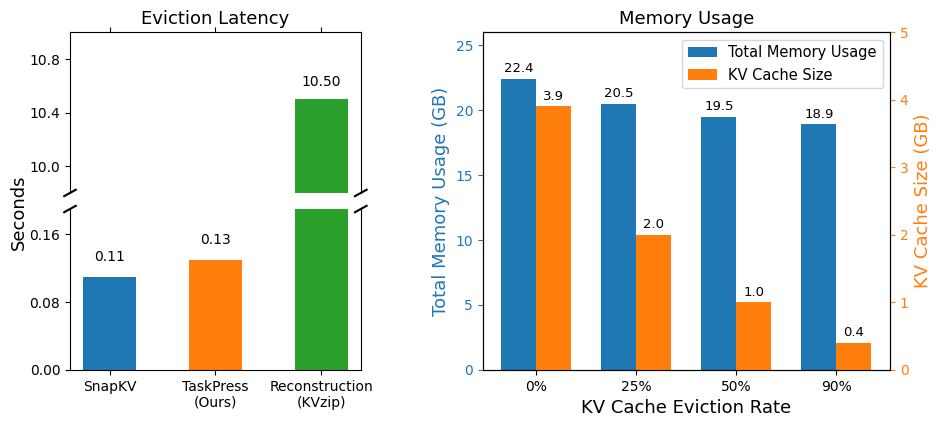}
    \vspace{-7mm}
    \caption{(Left) the eviction latency of each method. (Right) panel depicts the memory savings achieved during decoding after eviction. The experiments were conducted in a context length of 32K on LLaMA-3.1-8B.}
    \label{fig:eviction_perf}
    \vspace{-5mm}
\end{figure}

We analyzed the computational overhead of the eviction process.
Our method utilize attention map between context and task guides and eviction process has $O(L_t L_c)$ complexity. In contrast, reconstruction-based baselines \cite{kim2025kvzip} scale quadratically $O(L_c^2)$. Since $L_t \ll L_c$, our approach significantly lowers overhead, achieving an 80$\times$ latency improvement (see Fig. \ref{fig:eviction_perf}).

\subsubsection{Cost of task guide evolution}

% Using the setup in Section \ref{exp:setup} with 50 calibration samples, the entire procedure required a total runtime of 1,513.41 seconds (approximately 25.2 minutes). Breaking this down, the process required 5 external API calls (Gemini-2.0-Flash) for task guide initialization and mutation, taking 41.36 seconds and consuming 5,980 input and 486 output tokens. The majority of runtime was spent evaluating the calibration set via local inference, which took 1,459.71 seconds on a single A100 40GB GPU. However, we emphasize that this total runtime of ~25 minutes is a strictly one-time, offline setup cost. 

Using the setup in Section \ref{exp:setup} with 50 calibration samples, the entire procedure took $\sim$25.2 minutes. This comprised 41.4 seconds for 5 Gemini-2.0-Flash API calls (guide initialization and mutation; 5,980 input and 486 output tokens) and 1,459.7 seconds for local calibration set evaluation on a single A100 40GB GPU. Crucially, this $\sim$25-minute runtime is a strictly one-time, offline setup cost.

\subsubsection{Decoding and prefill performance}

\begin{table}[ht]
\vspace{-0.3cm}
\centering
\footnotesize 
\setlength{\tabcolsep}{2pt} 
\renewcommand{\arraystretch}{0.85} 
\caption{Performance impact during decoding.}
\begin{tabular}{l c c c}
\toprule
& \multicolumn{3}{c}{\textbf{Eviction Rate}} \\
\cmidrule(lr){2-4}
\textbf{Metric} & \textbf{0\%} & \textbf{50\%} & \textbf{75\%} \\
\midrule
\textbf{TPOT} (ms/token)         & 50.1 & 34.3 {\scriptsize(\textit{32\% $\downarrow$})} & 32.8 {\scriptsize(\textit{35\% $\downarrow$})} \\
\textbf{Throughput} (tokens/sec) & 19.9 & 29.2  {\scriptsize (\textit{46\% $\uparrow$})}   & 30.5 {\scriptsize(\textit{53\% $\uparrow$})} \\
\bottomrule
\end{tabular}
\label{tab:performance_metrics}
\end{table}

\begin{table}[ht]
\vspace{-0.2cm}
\centering
\small
\caption{Performance impact during the prefill phase. The value after the slash (/) indicates the total prompt length, including the task guide.}
\begin{tabular}{cccc}
\toprule
\textbf{Task Guide Length} & \textbf{TTFT (s)} & \textbf{Peak Memory (GB)} \\
\midrule
0 / 65536 (Baseline)  & 9.66 & 38.55 \\
64 / 65600        & 9.78 {\scriptsize (+1.2\%)} & 32.72 {\scriptsize (-15.1\%)} \\
256 / 65856        & 9.72 {\scriptsize (+0.6\%)} & 32.76 {\scriptsize (-15.0\%)} \\
1024 / 66880       & 9.87 {\scriptsize (+2.1\%)} & 32.98 {\scriptsize (-14.4\%)} \\
\bottomrule
\end{tabular}
\label{tab:task_guide_vertical}
\vspace{-5mm}
\end{table}

We evaluated performance impacts during the prefill and decoding phases---specifically time-to-first-token (TTFT), time-per-output-token (TPOT), throughput, and peak memory. Tests were conducted in a long context (64K) setup with LLaMA-3.1-8B on single A100 80GB. We adopted flash attention \cite{dao2023flashattention}.  Reducing the KV cache size directly accelerates decoding speed. While task-guided pruning often raises concerns about prefill overhead, our results show it is negligible. At a 75\% eviction rate with a 64-token task guide, TTFT increases by only 1.2\%. This is because TaskPress maintains linear complexity by computing attention strictly between the short task guide and the context. This initial overhead is easily offset by the resulting gains in decoding speed.

\subsubsection{Efficiency gain in multi-query setting}

Lastly, we analyzed the efficiency gains of KV cache reuse in a multi-query setting. Query-dependent methods inherently face a \textit{\textbf{memory-latency trade-off}}: because their eviction mechanisms rely on the specific query, systems are forced into one of two suboptimal strategies (see Figure~\ref{fig:eviction_strategies}).

% --- 통합된 Figure 시작 ---
\begin{figure}
\centering

% Strategy A (Red)
\textbf{\small Query-Dependent: Optimize Latency (Sacrifice Memory)}
\begin{lstlisting}[language=Python, basicstyle=\ttfamily\tiny, commentstyle=\color{gray}\itshape, keywordstyle=\color{blue}\bfseries, frame=single, xleftmargin=1em, breaklines=true, backgroundcolor=\color{red!5}, rulecolor=\color{red!50}]
full_kv = prefill(context)      # [One-time] Retain 100% KV
for q in queries:
    comp_kv = evict(full_kv, q) # [Per-query] Eviction
    generate(comp_kv, q)
\end{lstlisting}

\vspace{-0.5em}

% Strategy B (Orange)
\textbf{\small Query-Dependent: Optimize Memory (Sacrifice Latency)}
\begin{lstlisting}[language=Python, basicstyle=\ttfamily\tiny, commentstyle=\color{gray}\itshape, keywordstyle=\color{blue}\bfseries, frame=single, xleftmargin=1em, breaklines=true, backgroundcolor=\color{orange!5}, rulecolor=\color{orange!50}]
for q in queries:
    full_kv = prefill(context)  # [Per-query] Prefill
    comp_kv = evict(full_kv, q) # [Per-query] Eviction
    generate(comp_kv, q)
\end{lstlisting}

\vspace{-0.5em}

% TaskPress (Green)
\textbf{\small TaskPress: Query-Agnostic (Ours)}
\begin{lstlisting}[language=Python, basicstyle=\ttfamily\tiny, commentstyle=\color{gray}\itshape, keywordstyle=\color{blue}\bfseries, frame=single, xleftmargin=1em, breaklines=true, backgroundcolor=\color{green!5}, rulecolor=\color{green!60!black}]
full_kv = prefill(context)
comp_kv = evict(full_kv, task)  # [One-time] Compressed KV
for q in queries:
    generate(comp_kv, q)
\end{lstlisting}

\vspace{-0.5em}
\caption{Pseudo-code comparison of KV cache eviction strategies between query dependent and TaskPress.}
\label{fig:eviction_strategies}
\end{figure}
% --- 통합된 Figure 끝 ---

% The first strategy prioritizes latency at the expense of memory (Figure~\ref{fig:eviction_strategies}, top). To accommodate different compressions for future queries, the system permanently retains the full, uncompressed KV cache. This severely inflates the memory footprint, negating the purpose of compression. The second strategy prioritizes memory but sacrifices latency (Figure~\ref{fig:eviction_strategies}, middle). By discarding the full KV cache after generation, every new query forces a redundant full-context prefill from scratch to compute query-aware attention scores.

% TaskPress avoids this trade-off by decoupling context compression from the query (Figure~\ref{fig:eviction_strategies}, bottom). The only overhead is a strictly one-time prefill of a short task guide and a single eviction step. To empirically validate this, we benchmarked end-to-end latency and peak memory consumption across a varying number of queries on a single 8K-token context, applying a 75\% KV cache eviction.

The first strategy prioritizes latency by retaining the full KV cache, which severely inflates memory overhead and negates compression (Figure~\ref{fig:eviction_strategies}, top). Conversely, the second strategy prioritizes memory by discarding the cache, but suffers high latency since every new query forces a redundant full-context prefill (Figure~\ref{fig:eviction_strategies}, middle).

TaskPress breaks this trade-off by decoupling compression from the query (Figure~\ref{fig:eviction_strategies}, bottom), incurring only a one-time task guide prefill and eviction step. We validated this by benchmarking end-to-end latency and peak memory across varying query counts on an 8K-token context with 75\% eviction in Table \ref{tab:multi_query_latency}.
TaskPress outperforms both strategies.
Note that query-dependent's latency optimized strategy consumes 1,184.4 (MB) for KV cache which is \textit{5 times bigger} than TaskPress (236.88 MB). This is because it holds both compressed and full KV cache during generate.  

\begin{table}[h]
\centering
\caption{Comparison of end-to-end latency (sec) in a multi-query setting. Lat-Opt refers to latency optimized whereas Mem-Opt refers to memory optimized strategy.}
\footnotesize % 폰트 크기 축소
\setlength{\tabcolsep}{2pt} % 열 사이 간격 좁히기 (기본값 6pt)
\begin{tabular}{l c c c c c}
\toprule
 & \multicolumn{2}{c}{\textbf{Query-Dependent}} & \textbf{Ours} & \multicolumn{2}{c}{\textbf{E2E Gain (sec)}} \\
\cmidrule(lr){2-3} \cmidrule(lr){5-6}
\textbf{\# Queries} & \textbf{Lat-Opt} & \textbf{Mem-Opt} &  & \textbf{Lat-Opt} & \textbf{Mem-Opt} \\
\midrule
5   & 3.15  & 5.75   & 3.20  & -0.05 & 2.55 \\
10  & 5.62  & 11.52  & 5.56  & 0.06  & 5.96 \\
50  & 25.62 & 57.45  & 24.55 & 1.07  & 32.90 \\
200 & 50.24 & 115.39 & 48.06 & 2.18  & 67.33 \\
\bottomrule
\label{tab:multi_query_latency}
\end{tabular}
\vspace{-0.5cm}
\end{table}

\subsection{Analysis on Importance Scoring}
\subsubsection{Impacts of Key and Value score}
\label{sec:key_value_impact}
We analyzed impacts of the task guide score and value scale score using the single-document QA datasets. As shown in Table \ref{tab:eviction_results}, utilizing the task guide score alone yields better performance than the query-agnostic SnapKV baseline. Furthermore, combining the task guide score with value scale scoring results in the highest overall accuracy.

% Preamble required: \usepackage{booktabs}

\begin{table}[ht]
\centering
\caption{Key and Value importance score ablation study.}
\vspace{-2mm}
\resizebox{\columnwidth}{!}{%
\begin{tabular}{lcccc}
\toprule
& \multicolumn{2}{c}{Eviction ratio: 50\%} & \multicolumn{2}{c}{Eviction ratio: 75\%} \\
\cmidrule(lr){2-3} \cmidrule(lr){4-5}
 & Qasper & MF-en & Qasper & MF-en \\
\midrule
SnapKV /wo query & 40.64 & 46.36 & 30.27 & 35.96 \\
Task guide score & 41.08 & 52.49 & \textbf{33.66} & 42.93 \\
Task guide + Value outlier score & \textbf{42.12} & \textbf{54.78} & 32.29 & \textbf{45.76} \\
\bottomrule
\end{tabular}
}
\label{tab:eviction_results}
\vspace{-5mm}
\end{table}

\subsubsection{Scale factor as a proxy for Value norm}
\label{sec:value_score_justification}

Self-attention computes a weighted sum of value vectors, value magnitude is a critical indicator of token importance \cite{guo2024attention,devoto2025expected}. Building on this, our key insight is to utilize the scale factor as an efficient proxy for the value norm. This is supported by a strong Pearson correlation of $0.79 \pm 0.10$ between the two metrics. Furthermore, the scale factor shows near-zero correlation ($0.06 \pm 0.23$) with the attention map, indicating it captures orthogonal information. As detailed in Section \ref{sec:key_value_impact}, combining these complementary signals ultimately improves accuracy.

\subsection{Score aggregation strategy}

We compared three strategies to blend  $s^\text{key}$ and $s^\text{value}$: multiplication, addition with normalization, and the harmonic mean. Multiplication achieves the highest accuracy on both datasets (see Table~\ref{tab:blend_result}).

\begin{table}[ht]
\centering
\caption{Key and Value score unification ablation study (Eviction ratio: 75\%).}
\vspace{-2mm}
\resizebox{\columnwidth}{!}{%
\begin{tabular}{lcc}
\toprule
 & RULER 4K & MF-en \\
\midrule
Multiply $s^\text{key}\cdot s^{\text{value}}$ & 65.81 & 45.76 \\
Normalized Add $\frac{s^\text{key}}{\| \mathbf{s}^\text{key} \|_2} + \frac{s^\text{value}}{\| \mathbf{s}^\text{value} \|_2}$ & 59.90 & 43.28 \\
Harmonic Mean $\frac{2 s^\text{key} s^\text{value}}{s^\text{key}, s^\text{value}}$ & 64.14 & 42.66 \\
\bottomrule
\end{tabular}
}
\label{tab:blend_result}
\vspace{-5mm}
\end{table}

% \begin{table}[ht]
% \centering
% \caption{Key and Value score unification ablation study.}
% \vspace{-2mm}
% \resizebox{\columnwidth}{!}{%
% \begin{tabular}{lcccc}
% \toprule
% & \multicolumn{2}{c}{Eviction ratio: 50\%} & \multicolumn{2}{c}{Eviction ratio: 75\%} \\
% \cmidrule(lr){2-3} \cmidrule(lr){4-5}
%  & RULER 4K & MF-en & RULER 4K & MF-en \\
% \midrule
% Multiply $s^\text{key}\cdot s^{\text{value}}$ & - & 54.78 & 65.81	 & 45.76 \\
% Normalized Add $s^\text{key}+ s^{\text{value}}$ & - & 54.86 & 59.90	 & 43.28 \\
% Harmonic Mean $\frac{2 s^\text{key} s^\text{value}}{s^\text{key}, s^\text{value}}$ & - & 51.93
%  & 64.14
% 	 & 42.66 \\
% \bottomrule
% \end{tabular}
% }
% \label{tab:eviction_results}
% \vspace{-5mm}
% \end{table}

\subsection{Analysis on Task Guide}
\subsubsection{Robustness of underlying model}

To evaluate the influence of the source model, we prompted different models to generate a task guide using the same instruction. Table~\ref{tab:source_model} indicate that all evaluated models produced effective task guides, highlighting the robustness of underlying model.

\begin{table}[htbp]
\centering
% Increased compact spacing adjustments
\small % Reduces overall font size (can also try \footnotesize)
\setlength{\tabcolsep}{4pt} % Reduced from 8pt to 4pt for tighter columns
\renewcommand{\arraystretch}{0.85} % Reduced from 0.95 for tighter rows
\caption{NarrativeQA accuracy of source models.}
\begin{tabular}{l rr}
\toprule
\multirow{2}{*}{\textbf{Task Guide Source Model}} & \multicolumn{2}{c}{\textbf{Eviction Rate (\%)}} \\
\cmidrule(lr){2-3}
 & \textbf{50\%} & \textbf{75\%} \\
\midrule
Gemini Pro        & 30.38 & 30.11 \\
Claude Sonnet 4.6 & 29.74 & 29.09 \\
ChatGPT           & 29.22 & 29.02 \\
\bottomrule
\end{tabular}
\label{tab:source_model}
\vspace{-3mm}
\end{table}

\subsubsection{Impact of granularity or length}

 To control the specificity of the instructions, we vary the length of the task guide across three prompt variations. As shown in Table~\ref{tab:task_guide}, task guide with moderate length was optimal.

\begin{table}[htbp]
\centering
\small % Reduces overall font size
\setlength{\tabcolsep}{4pt} % Extremely tight horizontal spacing
\renewcommand{\arraystretch}{0.9} % Very tight vertical spacing
\caption{Impacts of task guide granularity on NarrativeQA accuracy.}
\vspace{-2mm}
\begin{tabular}{ll rr}
\toprule
\multirow{2}{*}{\textbf{Granularity}} & \textbf{Length} & \multicolumn{2}{c}{\textbf{Accuracy (\%)}} \\
\cmidrule(lr){3-4}
 & \textbf{(sentences)} & \textbf{50\%} & \textbf{75\%} \\
\midrule
Generic       & 1                & 29.08 & 28.32 \\
Specific      & 2--3             & 30.38 & 30.11 \\
Very Specific & \textgreater{} 5 & 29.93 & 27.67 \\
\bottomrule
\end{tabular}
\label{tab:task_guide}
\vspace{-2mm}
\end{table}

\subsubsection{Impact of paraphrasing}

We evaluated the impact of wording or paraphrasing. Firstly, we replaced five words in the task guide with synonyms and combined them to generate five distinct prompt (P) sets. Secondly, to measure the impact of sentence-level variations, we paraphrased the task guide into several alternative sentences. The overall robustness is consistently maintained with low (1\%<) standard deviation.

% Make sure to include these packages in your preamble:
% \usepackage{booktabs}
% \usepackage{multirow}
% \usepackage{graphicx}

\begin{table}[h]
\centering
\caption{Impacts of guide wording variations. `W' denotes word-level (synonym replacements) and `S' denotes sentence-level (paraphrasing).}
\vspace{-1mm}
\resizebox{\columnwidth}{!}{%
\begin{tabular}{llcccccc}
\toprule
\textbf{Dataset} & \textbf{Lvl} & \textbf{Average} & \textbf{P-1} & \textbf{P-2} & \textbf{P-3} & \textbf{P-4} & \textbf{P-5} \\
\midrule
\multirow{2}{*}{NarrativeQA} 
& W & 31.01$\pm$0.25 & 30.58 & 31.17 & 30.85 & 31.26 & 30.75 \\
& S & 30.41$\pm$0.59 & 30.58 & 30.10 & 30.52 & 31.21 & 29.63 \\
\midrule
\multirow{2}{*}{Qasper} 
& W & 43.99$\pm$0.76 & 45.53 & 44.41 & 43.17 & 43.55 & 44.82 \\
& S & 43.67$\pm$1.29 & 45.53 & 41.99 & 43.66 & 43.15 & 44.02 \\
\midrule
\multirow{2}{*}{MF-en} 
& W & 53.93$\pm$0.41 & 54.42 & 53.37 & 54.31 & 53.91 & 54.14 \\
& S & 54.14$\pm$0.38 & 54.42 & 54.41 & 54.38 & 53.95 & 53.56 \\
\bottomrule
\end{tabular}%
}
\label{tab:ablation}
\vspace{-3mm}
\end{table}

\subsubsection{Task drift during conversation}

To assess the impact of misaligned task guides on model performance, we conducted experiments under two conditions: using guides from (1) similar but non-target QA tasks with different semantic focuses, and (2) entirely disparate tasks (e.g., NarrativeQA and RULER's needle-in-a-haystack). As shown in Table \ref{tab:applied_task_guide_moderate}, the original task-aligned guide yields the highest accuracy. Notably, performance degradation becomes significant with highly dissimilar guides and higher eviction rates.

\begin{table}[htbp]
\centering
\small % 폰트 크기 축소
\setlength{\tabcolsep}{4pt} % 타이트한 열 간격
\renewcommand{\arraystretch}{0.9} % 타이트한 행 간격
\caption{Impacts of misaligned task guide on accuracy.}

\begin{tabular}{ll rr}
\toprule
\multirow{2}{*}{\textbf{Task Guide Source}} & \multirow{2}{*}{\textbf{Evaluated Task}} & \multicolumn{2}{c}{\textbf{Accuracy (\%)}} \\
\cmidrule(lr){3-4}
 & & \textbf{50\%} & \textbf{75\%} \\
\midrule
\textbf{NarrativeQA (Correct)} &  NarrativeQA & \textbf{30.38} & \textbf{30.11} \\
HotpotQA                       & NarrativeQA &  29.72 & 28.31 \\
2WikiMultiHopQA                & NarrativeQA &  29.70 & 27.86 \\
MuSiQue                        & NarrativeQA &  29.30 & 27.95 \\
DuReader                       & NarrativeQA &  29.05 & 28.00 \\
\hline \\
\textbf{RULER 4K (Correct)} & RULER 4K & - & \textbf{62.2} \\
NarrativeQA & RULER 4K & - &  58.7 \\
\bottomrule
\end{tabular}
\label{tab:applied_task_guide_moderate}
\vspace{-2mm}
\end{table}

\section{Conclusion}

We present TaskPress, a framework that achieves efficient, query-agnostic KV cache compression by utilizing task guides as semantic anchors and repurposing quantization scale factors for outlier detection, thereby significantly outperforming existing baselines on standard long-context benchmarks.
\section{Limitations}

Despite the effectiveness of TaskPress in balancing efficiency and flexibility, few limitations remain.

\noindent\textbf{Dependency on Task Scope.} While our approach is query-agnostic within a specific domain, it remains task-dependent. The compressed KV cache is optimized strictly for the semantic scope defined by the task guide. Consequently, if user queries drift significantly from the anticipated task (e.g., asking a coding question during a document summarization session), the retrieved context may be insufficient, potentially degrading performance compared to full-cache baselines.

\noindent\textbf{Extremely Dense Information.} A limitation of TaskPress is its vulnerability to severe attention dilution in artificially dense environments, such as the "needle inside needles" scenario in RULER MK-2 and MK-3. Because we use softmax-normalized cross-attention to score token importance, a context entirely saturated with target-like entities scatters the attention mass. This causes individual target scores to drop below background noise, leading to unintended cache eviction. However, such extreme density rarely reflects real-world workloads. As detailed in Appendix \ref{app:mk23}, TaskPress maintains robust retrieval performance provided the targets appear at a moderate, realistic frequency.

% \noindent\textbf{Reliance on Quantization Statistics.} Our outlier detection leverages scale factors inherent to quantized models as a zero-cost proxy for importance. In full-precision settings (e.g., FP16 or BF16) where these pre-computed artifacts are absent, the requisite statistics must be derived explicitly. However, this imposes negligible overhead, as the equivalent metric the maximum absolute value per token can be computed efficiently on-the-fly.

\noindent\textbf{Reliance on Quantization Statistics.} Our outlier detection leverages scale factors inherent to quantized models as a zero-cost proxy for importance. In full-precision settings (e.g., FP16 or BF16) where these pre-computed artifacts are absent, the requisite statistics must be derived explicitly. However, this imposes negligible overhead, as the equivalent metric—the maximum absolute value per token—can be computed efficiently on-the-fly. For instance, processing a 64K token context using LLaMA-3.1-8B in full precision requires 9.66 seconds for the overall prefill, whereas dynamically calculating the absolute min-max adds merely 0.26 seconds. This represents a marginal overhead of roughly 2.7%, ensuring the core approach remains highly efficient and applicable even in non-quantized serving scenarios.

\bibliography{custom}

@article{comanici2025gemini,
  title={Gemini 2.5: Pushing the frontier with advanced reasoning, multimodality, long context, and next generation agentic capabilities},
  author={Comanici, Gheorghe and Bieber, Eric and Schaekermann, Mike and Pasupat, Ice and Sachdeva, Noveen and Dhillon, Inderjit and Blistein, Marcel and Ram, Ori and Zhang, Dan and Rosen, Evan and others},
  journal={arXiv preprint arXiv:2507.06261},
  year={2025}
}

@article{dubey2024llama,
  title={The llama 3 herd of models},
  author={Dubey, Abhimanyu and Jauhri, Abhinav and Pandey, Abhinav and Kadian, Abhishek and Al-Dahle, Ahmad and Letman, Aiesha and Mathur, Akhil and Schelten, Alan and Yang, Amy and Fan, Angela and others},
  journal={arXiv e-prints},
  pages={arXiv--2407},
  year={2024}
}

@article{li2024snapkv,
  title={Snapkv: Llm knows what you are looking for before generation},
  author={Li, Yuhong and Huang, Yingbing and Yang, Bowen and Venkitesh, Bharat and Locatelli, Acyr and Ye, Hanchen and Cai, Tianle and Lewis, Patrick and Chen, Deming},
  journal={Advances in Neural Information Processing Systems},
  volume={37},
  pages={22947--22970},
  year={2024}
}

@article{kim2025kvzip,
  title={KVzip: Query-Agnostic KV Cache Compression with Context Reconstruction},
  author={Kim, Jang-Hyun and Kim, Jinuk and Kwon, Sangwoo and Lee, Jae W and Yun, Sangdoo and Song, Hyun Oh},
  journal={arXiv preprint arXiv:2505.23416},
  year={2025}
}

@article{zhang2023h2o,
  title={H2o: Heavy-hitter oracle for efficient generative inference of large language models},
  author={Zhang, Zhenyu and Sheng, Ying and Zhou, Tianyi and Chen, Tianlong and Zheng, Lianmin and Cai, Ruisi and Song, Zhao and Tian, Yuandong and R{\'e}, Christopher and Barrett, Clark and others},
  journal={Advances in Neural Information Processing Systems},
  volume={36},
  pages={34661--34710},
  year={2023}
}

@inproceedings{bai2024longbench,
  title={Longbench: A bilingual, multitask benchmark for long context understanding},
  author={Bai, Yushi and Lv, Xin and Zhang, Jiajie and Lyu, Hongchang and Tang, Jiankai and Huang, Zhidian and Du, Zhengxiao and Liu, Xiao and Zeng, Aohan and Hou, Lei and others},
  booktitle={Proceedings of the 62nd annual meeting of the association for computational linguistics (volume 1: Long papers)},
  pages={3119--3137},
  year={2024}
}

@article{hsieh2024ruler,
  title={RULER: What's the Real Context Size of Your Long-Context Language Models?},
  author={Cheng-Ping Hsieh and Simeng Sun and Samuel Kriman and Shantanu Acharya and Dima Rekesh and Fei Jia and Yang Zhang and Boris Ginsburg},
  year={2024},
  journal={arXiv preprint arXiv:2404.06654},
}

@article{yang2025qwen3,
  title={Qwen3 technical report},
  author={Yang, An and Li, Anfeng and Yang, Baosong and Zhang, Beichen and Hui, Binyuan and Zheng, Bo and Yu, Bowen and Gao, Chang and Huang, Chengen and Lv, Chenxu and others},
  journal={arXiv preprint arXiv:2505.09388},
  year={2025}
}

@article{liu2024kivi,
  title={Kivi: A tuning-free asymmetric 2bit quantization for kv cache},
  author={Liu, Zirui and Yuan, Jiayi and Jin, Hongye and Zhong, Shaochen and Xu, Zhaozhuo and Braverman, Vladimir and Chen, Beidi and Hu, Xia},
  journal={arXiv preprint arXiv:2402.02750},
  year={2024}
}

@article{kovcisky2018narrativeqa,
  title={The narrativeqa reading comprehension challenge},
  author={Ko{\v{c}}isk{\`y}, Tom{\'a}{\v{s}} and Schwarz, Jonathan and Blunsom, Phil and Dyer, Chris and Hermann, Karl Moritz and Melis, G{\'a}bor and Grefenstette, Edward},
  journal={Transactions of the Association for Computational Linguistics},
  volume={6},
  pages={317--328},
  year={2018},
  publisher={MIT Press One Rogers Street, Cambridge, MA 02142-1209, USA journals-info~…}
}

@inproceedings{zhong2021qmsum,
  title={QMSum: A new benchmark for query-based multi-domain meeting summarization},
  author={Zhong, Ming and Yin, Da and Yu, Tao and Zaidi, Ahmad and Mutuma, Mutethia and Jha, Rahul and Hassan, Ahmed and Celikyilmaz, Asli and Liu, Yang and Qiu, Xipeng and others},
  booktitle={Proceedings of the 2021 Conference of the North American Chapter of the Association for Computational Linguistics: Human Language Technologies},
  pages={5905--5921},
  year={2021}
}

@article{dao2023flashattention,
  title={Flashattention-2: Faster attention with better parallelism and work partitioning},
  author={Dao, Tri},
  journal={arXiv preprint arXiv:2307.08691},
  year={2023}
}

@article{xiao2023efficient,
  title={Efficient Streaming Language Models with Attention Sinks},
  author={Xiao, Guangxuan and Tian, Yuandong and Chen, Beidi and Han, Song and Lewis, Mike},
  journal={arXiv preprint arXiv:2309.17453},
  year={2023}
}

@article{cai2024pyramidkv,
  title={{PyramidKV}: Dynamic KV Cache Compression based on Pyramidal Information Funneling},
  author={Cai, Zefan and Zhang, Yichi and Gao, Bofei and Liu, Yuliang and Li, Yucheng and Liu, Tianyu and Lu, Keming and Xiong, Wayne and Dong, Yue and Hu, Junjie and others},
  journal={arXiv preprint arXiv:2406.02069},
  year={2024}
}

@article{tang2024quest,
  title={Quest: Query-Aware Sparsity for Efficient Long-Context LLM Inference},
  author={Tang, Jiaming and Zhao, Yilong and Zhu, Kan and Xiao, Guangxuan and Kasikci, Baris and Han, Song},
  journal={arXiv preprint arXiv:2406.10774},
  year={2024}
}

@article{devoto2025expected,
  title={Expected Attention: KV Cache Compression by Estimating Attention from Future Queries Distribution},
  author={Devoto, Alessio and Jeblick, Maximilian and J{\'e}gou, Simon},
  journal={arXiv preprint arXiv:2510.00636},
  year={2025}
}

@article{liu2024spinquant,
  title={SpinQuant: LLM quantization with learned rotations},
  author={Liu, Zechun and Zhao, Changsheng and Fedorov, Igor and Soran, Bilge and Choudhary, Dhruv and Krishnamoorthi, Raghuraman and Chandra, Vikas and Tian, Yuandong and Blankevoort, Tijmen},
  journal={arXiv preprint arXiv:2405.16406},
  year={2024}
}

@article{sengupta2025value,
  title={Value-Guided KV Compression for LLMs via Approximated CUR Decomposition},
  author={Sengupta, Ayan and Chaudhary, Siddhant and Chakraborty, Tanmoy},
  journal={arXiv preprint arXiv:2509.15038},
  year={2025}
}

@article{agrawal2025gepa,
  title={{GEPA}: Reflective prompt evolution can outperform reinforcement learning},
  author={Agrawal, Lakshya A and Tan, Shangyin and Soylu, Dilara and Ziems, Noah and Khare, Rishi and Opsahl-Ong, Krista and Singhvi, Arnav and Shandilya, Herumb and Ryan, Michael J and Jiang, Meng and others},
  journal={arXiv preprint arXiv:2507.19457},
  year={2025}
}

@inproceedings{xu2025think,
  title={Think: Thinner key cache by query-driven pruning},
  author={Xu, Yuhui and Jie, Zhanming and Dong, Hanze and Wang, Lei and Lu, Xudong and Zhou, Aojun and Saha, Amrita and Xiong, Caiming and Sahoo, Doyen},
  booktitle={International Conference on Learning Representations},
  volume={2025},
  pages={56691--56709},
  year={2025}
}

@inproceedings{guo2024attention,
  title={Attention score is not all you need for token importance indicator in kv cache reduction: Value also matters},
  author={Guo, Zhiyu and Kamigaito, Hidetaka and Watanabe, Taro},
  booktitle={Proceedings of the 2024 Conference on Empirical Methods in Natural Language Processing},
  pages={21158--21166},
  year={2024}
}

\appendix

\newpage
\newpage

\section{Task Guides used in our experiments}

Table \ref{tab:combined_qa_sum} details the task guides employed in our LongBench \cite{bai2024longbench} experiments for single- and multi-document question answering. These guides were primarily generated by Gemini, with minimal human refinement.

Table \ref{tab:synthetic_stress_test} details the task guide employed for our evaluation on the RULER benchmark \cite{hsieh2024ruler}. RULER comprises a diverse set of sub-tasks, including long-context retrieval (Needle-in-a-Haystack), multi-hop tracing, aggregation, and question answering. Notably, we utilize a single, unified task guide across all sub-tasks, demonstrating that our method effectively generalizes across heterogeneous tasks without requiring task-specific customization.

The task guides used in other tasks are provided in our attached source code.

\section{Explanation of the MK-2/3 lower accuracy in RULER}
\label{app:mk23}

In this section, we explain why TaskPress has relatively lower accuracy on MK-2 and MK-3 tasks of RULER.
MK-2/3 are the tasks that require finding “a needle inside needles”. Below is an example of context and question drawn from MK-2. MK-3 shares the same schematic except the key and value are UUIDs.

\begin{tcolorbox}[colback=gray!15, colframe=gray!15, sharp corners, fontupper=\tiny\ttfamily]
\begin{verbatim}
Context: A special magic number is hidden within the following text. 
Make sure to memorize it.
...
One of the special magic numbers for confused-prince is: 4446065.  
One of the special magic numbers for innocent-clause is: 5774152.  
....  
One of the special magic numbers for woebegone-emission is: 4123902.  

Question: What is the special magic number for innocent-clause 
mentioned in the provided text?
\end{verbatim}
\end{tcolorbox}

% Our task guide instructed the model to capture a ``needle'' of magic numbers or UUIDs. TaskPress calculates KV cache importance by taking the mean cross-attention scores between the task guide and the entire context. However, because the context is full of needles ($> 820$), attention scores are spread thin, causing attention dilution.

% In MK-1, the task guide sharply focuses on a single needle, assigning the target key an attention score of $3.01\times 10^{-4}$, which is roughly six times higher than the context mean. However, MK-2 and MK-3 introduce multiple needles, forcing the attention mass to scatter. Consequently, the attention score for any single target key drastically plummets to $1.2 \times 10^{-5}$---falling significantly below the mean background noise of $5.0 \times 10^{-5}$.

% Since it is highly unrealistic for real-world data to consist exclusively of target "needles," we conducted an experiment by incrementally increasing the number of needles inserted into the context. Using the official repository, we re-generated the RULER MK-2 dataset with varying needle counts. Our results demonstrate that as long as the context contains a moderate, realistic number of needles—rather than an overwhelming density—TaskPress maintains robust retrieval performance and effectively captures the targets under presence of multiple needles.

Our task guide instructed the model to capture a ``needle" with the magic numbers or UUIDs, e.g., a snippet of task guide is \textit{"...long-context retrieval by hiding specific facts, such as arbitrary magic numbers or UUIDs in key-value forms..."}.

TaskPress calculates KV cache importance by taking the mean cross-attention scores between the task guide and the entire context. However, because the context is full of needles ($> 820$), attention scores are spread thin, causing attention dilution.

We conducted an empirical measurement of attention mass of RULER-16K. In MK-1, the task guide sharply focuses on a single needle, assigning the target key an attention score of $3.01\times 10^{-4}$, which is roughly six times higher than the context mean. However, MK-2 and MK-3 introduce multiple needles, forcing the attention mass to scatter. Consequently, the attention score for any single target key drastically plummets to $1.2 \times 10^{-5}$---falling significantly below the mean background noise of $5.0 \times 10^{-5}$.

Since it is highly unrealistic for real-world data to consist exclusively of target "needles," we conducted an experiment by incrementally increasing the number of needles inserted into the context. Using the official repository, we re-generated the RULER MK-2 dataset with varying needle counts. Our results demonstrate that as long as the context contains a moderate, realistic number of needles—rather than an overwhelming density—TaskPress maintains robust retrieval performance and effectively captures the targets under presence of multiple needles.

\begin{table}[htbp]
    \centering
    \caption{RULER-16K MK-2 results varying the number of needles}
    \label{tab:ruler_mk2_needles}
    \begin{tabular}{ccc}
        \toprule
        \textbf{\# Needles} & \textbf{50\% Eviction} & \textbf{75\% Eviction} \\
        \midrule
        1  & 100.0 & 99.0 \\
        5  & 99.8  & 96.0 \\
        10 & 98.8  & 93.4 \\
        25 & 98.0  & 74.8 \\
        \bottomrule
    \end{tabular}
\end{table}
\begin{table*}[b]
    \centering
    \small
    \caption{Task guides used in our LongBench QA experiments and randomly sampled example queries for each task.}
    \renewcommand{\arraystretch}{1.5}
    \begin{tabularx}{\textwidth}{l p{0.35\textwidth} X}
        \toprule
        \textbf{Dataset} & \textbf{Task Guide (Meta-Query)} & \textbf{Queries (Examples)} \\
        \midrule
        \multicolumn{3}{c}{\textit{Single-Document QA}} \\
        \midrule
        \textbf{NarrativeQA} & 
        Single-Document QA task based on long stories and scripts, such as books or movie screenplays. The model is required to answer specific questions that demand a deep understanding of the narrative flow, character interactions, and plot development over the entire lengthy context. & 
        Q: What town did Daisy encounter Jim in? \newline 
        Q: What is unique about the phonetic spelling of the 'future' article? \\
        \midrule
        \textbf{Qasper} & 
        Single-Document QA task focused on extracting information from Natural Language Processing (NLP) research papers. The model must locate and interpret specific details, experimental results, or methodological descriptions within a dense scientific text. & 
        Q: What is the GhostVLAD approach? \newline 
        Q: How does Gaussian-masked directional multi-head attention work? \\
        \midrule
        \textbf{MultiFieldQA-en} & 
        Single-Document QA task that tests the model's ability to handle long contexts across a diverse range of fields, including law, government reports, and technical manuals, in English. This evaluates the model's adaptability to specific domain vocabularies and varied document structures. & 
        Q: How is the vacuum processing system configured? \newline 
        Q: When did the London Paving and Lighting Act pass? \\
        \midrule
        \multicolumn{3}{c}{\textit{Multi-Document QA}} \\
        \midrule
        \textbf{HotpotQA} & 
        Multi-Document QA task involving multi-hop reasoning, where the answer cannot be found in a single sentence or document. Multiple retrieved documents are concatenated, and the model must bridge information to derive the correct answer & 
        Q: When was the American singer... whose second studio album is Chapter II born? \newline 
        Q: Where did the punter for the Dallas Cowboys in the 1980s play college football? \\
        \midrule
        \textbf{2WikiMQA} & 
        Multi-Document QA task reasoning across disjoint passages to derive an answer. It involves a logical chain of entities and relationships that spans multiple paragraphs to reach the final conclusion. Focus on specific segments that establish the semantic relationships across disjoint passages. & 
        Q: What is the place of birth of the director of film A Chrysanthemum Bursts In Cincoesquinas? \newline 
        Q: Who died first, Erich Haenisch or William Pooley? \\
        \midrule
        \textbf{Musique} & 
        Multi-Document QA designed to minimize shortcuts. Demands rigorous, multi-step reasoning across unconnected texts to synthesize the answer. & 
        Q: What is the source of the river Orlam clans crossed to migrate to Namibia? \newline 
        Q: In what region of S-Fone is the place of birth of John Phan located? \\
        % \midrule
        % \multicolumn{3}{c}{\textit{Summarization}} \\
        % \midrule
        % \textbf{GovReport} & 
        % Condensing long, dense government reports into summaries. Captures policy objectives and statistical findings while ignoring filler. & 
        % Now, write a one-page summary of the report. \\
        % \midrule
        % \textbf{QMSum} & 
        % Query-based summarization of long meeting transcripts. The model extracts relevant parts from messy spoken language based on user queries. & 
        % Query: Summarize the whole meeting. \newline 
        % Query: What is the conclusion of the discussion about Marketing strategy? \\
        % \midrule
        % \textbf{MultiNews} & 
        % Synthesizing a single coherent news story from multiple articles. Requires resolving conflicting details and merging information. & 
        % Now, write a one-page summary of all the news. \\
        \bottomrule
    \end{tabularx}
    \label{tab:combined_qa_sum}
\end{table*}

\begin{table*}
    \centering
    \small
    \renewcommand{\arraystretch}{1.5}
    \caption{Task guide used in our RULER experiments and randomly sampled example queries.}
    \begin{tabularx}{\textwidth}{l p{0.35\textwidth} X}
        \toprule
        \textbf{Dataset} & \textbf{Task Guide (Meta-Query)} & \textbf{Queries (Examples)} \\
        \midrule
        \textbf{RULER} & 
        Evaluates long-context retrieval by hiding specific facts, such as arbitrary magic numbers or UUIDs its in key-value forms so remember both of them. Evaluate reasoning capabilities through coreference chain resolution of Variable (VAR) assignment. Additionally, frequent word extraction as a proxy for summarization. & 
        Q: What are all the special magic numbers for absorbed-pagoda mentioned in the text? \newline
        Q: What are the 10 most common words in the above list? \newline
        Q: What are all the special magic numbers for quick-campaign and spotless-undertaker? \\
        \bottomrule
    \end{tabularx}
    \label{tab:synthetic_stress_test}
\end{table*}

\section{Mutation \& Crossover Prompt}
\label{sec:mutation-crossover-prompt}

The evolutionary task-guide search in \texttt{evolve\_task\_guide.py}
uses a single large-language-model prompt to drive the variation step
of each generation. After the top-$k$ parent guides are selected by
fitness on the calibration set, the prompt in
``Mutation \& Crossover Prompt Template'' is sent to the LLM (Gemini in our setup)
together with the parents and a batch of real calibration queries.
The LLM is asked to return $N_\text{offspring}$ new candidate guides,
each produced by \emph{either} \textbf{mutation} (query-driven
rewriting of a single parent) \emph{or} \textbf{crossover}
(structure-preserving blending of two parents). The output guides
are then evaluated on the calibration set, and the cycle repeats for
the next generation.

The template is intentionally short and constrained
(2--3 sentences, ${<}64$ tokens per guide, no commentary) so that the
LLM produces drop-in replacements rather than free-form explanations.
The slot legend below lists which substrings are filled in at runtime
and which fragments are optional when the downstream task name is not
known a priori.

\definecolor{promptbg}{HTML}{F7F7F2}
\definecolor{promptrule}{HTML}{2F4F4F}
\definecolor{slot}{HTML}{B22222}
\definecolor{optional}{HTML}{1F6FEB}

\newcommand{\slot}[1]{\textcolor{slot}{\textsf{\{#1\}}}}
\newcommand{\opt}[1]{\textcolor{optional}{\textit{#1}}}

\begin{tcolorbox}[
  breakable,
  enhanced,
  colback=promptbg,
  colframe=promptrule,
  title=\textbf{Mutation \& Crossover Prompt Template},
  fonttitle=\bfseries,
  arc=2pt, boxrule=0.6pt
]
\small\sffamily
You are an expert at refining task guides for AI systems.\\[0.4em]

\opt{// Optional task-name header --- omit the entire line when the task identity
is unknown a priori.}\\
Current top-performing task guides\opt{[ for `\slot{task\_name}'\,]}:\\
\slot{parents\_text}\\[0.4em]

Here are real sample queries users might ask:\\
\slot{queries\_text}\\[0.4em]

Your goal is to produce \slot{n\_offspring} improved task guides.
For each guide, apply EITHER mutation OR crossover:\\[0.4em]

\textbf{Mutation (query-driven exploration \& refinement):}
\begin{itemize}\setlength\itemsep{0pt}
  \item Deeply analyze the sample queries to extract specific clues,
        constraints, and implicit intents required for retrieval
        (e.g.\ temporal limits, exact formatting, key entities).
  \item Explicitly embed these extracted query traits into the parent
        guides.
  \item While you can make targeted semantic edits, you must also
        occasionally make BOLD, SIGNIFICANT structural changes or
        entirely fresh rephrasings to explore new instruction spaces.
\end{itemize}

\textbf{Crossover (structure-preserving blending):}
\begin{itemize}\setlength\itemsep{0pt}
  \item When combining parent guides, preserve the core phrasing and
        structure of each parent. Blend their complementary ideas by
        splicing or interleaving specific clauses, rather than
        rewriting from scratch.
\end{itemize}

\medskip
Output exactly \slot{n\_offspring} distinct task guides, one per line,
numbered \textsf{1.} to \textsf{\slot{n\_offspring}.}
Each guide must be 2--3 sentences and less than 64 tokens.
No extra commentary.
\end{tcolorbox}

\paragraph{Slot legend.}
\textcolor{slot}{\textsf{\{red\}}} entries are runtime substitutions:
\slot{parents\_text} is the newline-joined list of parent guides,
\slot{queries\_text} is the newline-joined calibration queries, and
\slot{n\_offspring} is the requested offspring count.
\textcolor{optional}{\textit{Blue italic}} fragments mark the
\emph{optional} task-identity hint:
when the downstream task is \textbf{unknown a priori}
(e.g.\ open-domain deployment, ablation studies, or settings where
revealing the task name would bias the search),
both the comment line and the ``for `\slot{task\_name}'\,'' clause
should be removed so the LLM mutates the guides purely from the
calibration queries.

\end{document}